\pdfoutput=1

\documentclass[11pt]{article}

\usepackage[review]{acl}

\usepackage{times}
\usepackage{latexsym}

\usepackage[T1]{fontenc}

\usepackage[utf8]{inputenc}

\usepackage{microtype}

\usepackage{inconsolata}

\usepackage{amsmath,amsfonts,bm}

\def\eqref#1{equation~\ref{#1}}

\def\1{\bm{1}}

\DeclareMathAlphabet{\mathsfit}{\encodingdefault}{\sfdefault}{m}{sl}
\SetMathAlphabet{\mathsfit}{bold}{\encodingdefault}{\sfdefault}{bx}{n}

\usepackage{url}

\usepackage{graphicx}
\usepackage{hyperref}       
\usepackage{url}            
\usepackage{booktabs}       
\usepackage{amsfonts}       
\usepackage{nicefrac}       
\usepackage{microtype}      
\usepackage{xcolor}         
\usepackage{amssymb}
\usepackage{amsmath}
\usepackage{multirow}
\usepackage{makecell}
\usepackage{wrapfig}
\usepackage{subcaption}
\usepackage{bm}
\usepackage{bbding}
\usepackage{utfsym}
\usepackage{colortbl}
\newcommand{\modelname}{UniFashion}

\definecolor{mygray}{gray}{.9}

\title{
UniFashion: A Unified Vision-Language Model for Multimodal Fashion Retrieval and Generation
}

\author{ Xiangyu Zhao\textsuperscript{1}, Yuehan Zhang\textsuperscript{2}, Wenlong Zhang\textsuperscript{1,3}, Xiao-Ming Wu\textsuperscript{1}{\Envelope} \\ \textsuperscript{1}Department of Computing, The Hong Kong Polytechnic University\\  \textsuperscript{2}Wuhan University, \textsuperscript{3}Shanghai AI Laboratory \\xiang-yu.zhao@connect.polyu.hk, xiao-ming.wu@polyu.edu.hk}

\begin{document}

\maketitle



\begin{abstract}

The fashion domain includes a range of real-world multimodal tasks, such as multimodal retrieval and generation. Recent advancements in AI-generated content, particularly large language models for text and diffusion models for visuals, have spurred significant research interest in applying these multimodal models to fashion. However, fashion models must also effectively handle embedding tasks, like image-to-text and text-to-image retrieval. Moreover, current unified fashion models often lack the capability for image generation. In this work, we present \modelname{}, a unified framework that tackles the challenges of multimodal generation and retrieval tasks in the fashion domain, by integrating image and text generation with retrieval tasks. \modelname{} unifies embedding and generative processes through the use of a diffusion model and LLM, enabling controllable and high-fidelity generation. Our model significantly outperforms previous state-of-the-art models focused on single tasks across various fashion-related challenges and can be easily adapted to manage complex vision-language tasks. This study highlights the synergistic potential between multimodal generation and retrieval, offering a promising avenue for future research in the fashion domain. The source code is available at \url{https://github.com/xiangyu-mm/UniFashion}.


\end{abstract}

\section{Introduction}

The fashion domain presents a range of real-world multimodal tasks, encompassing multimodal retrieval~\citep{gao2020fashionbert, wu2021fashion, bai2023sentence, liu2024multimodal} and multimodal generation~\citep{yang2020fashion} tasks. Such tasks have been utilized in diverse e-commerce scenarios to enhance product discoverability, seller-buyer interaction, and customer conversion rates after catalog browsing~\citep{han2023fame, zhuge2021kaleido}. 
The remarkable progress in the field of artificial intelligence generated content (AIGC), particularly in technologies like large language models (LLMs)~\citep{vicuna2023,touvron2023llama,brown2020language} for text generation and diffusion models~\cite{rombach2022high, nichol2022glide,saharia2022photorealistic} for visual generation, yielding significant advancements in numerous downstream tasks~\citep{feng2023towards, zhang2022fine} and sparking widespread research interest in applying these multimodal models to the fashion domain.

Instruction-tuned multimodal large language models~\citep{liu2023visual, instructblip, dong2023dreamllm, zhao-etal-2024-easygen} (MLLMs) have emerged as a promising direction for developing a single multi-task model~\citep{shi2023recon}.
However, due to the heterogeneous nature of multimodal fashion tasks~\citep{han2023fame}, most existing MLLMs struggle to be directly applicable in the fashion domain. For example, in the fashion domain, retrieval tasks that rely on embedding ability, such as image-to-text or text-to-image retrieval, have largely been overlooked. 
Furthermore, existing MLLMs lack the ability to solve the composed image retrieval (CIR)~\citep{liu2021image, baldrati2022effective} task, which composes the reference image and related caption in a joint embedding to calculate similarities with candidate images and is particularly relevant in recommender systems~\citep{han2017automatic, liu2022boosting, liu2024vector}. 

Drawing inspiration from GRIT~\citep{muennighoff2024generative}, which successfully combined generative and embedding tasks into a unified model for text-centric applications and enhanced embedding performance by incorporating a generative objective, it is evident that exploring task correlations and integrating embedding with generative models in the fashion domain is promising.

\begin{figure*}[t]
	\centering
	\includegraphics[width=1.0\textwidth]{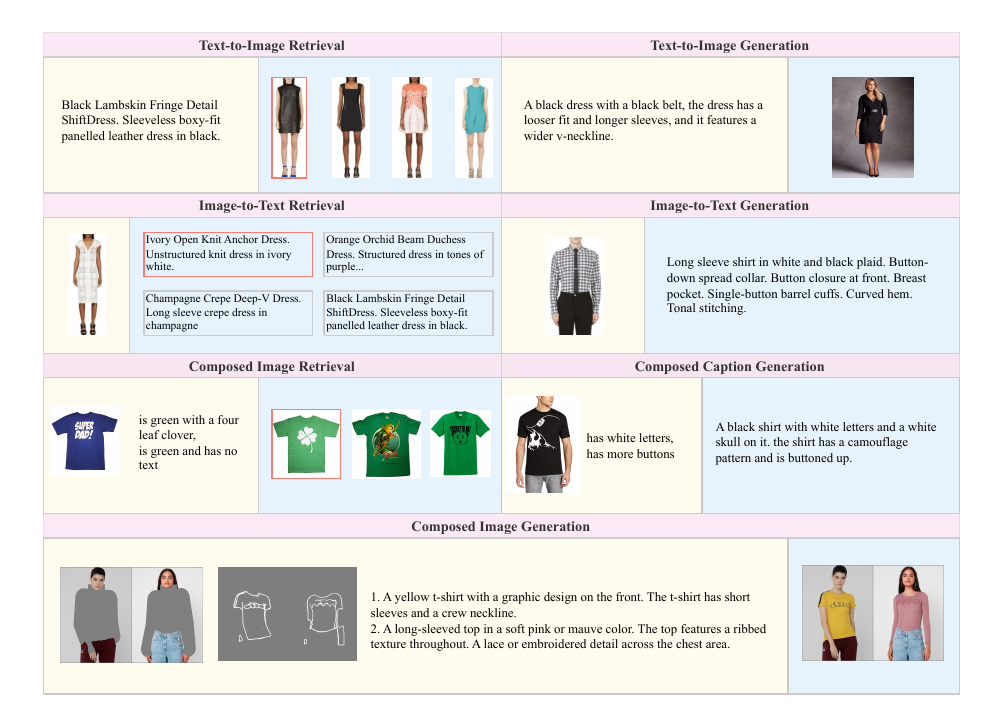}
        \caption{Illustration of the fashion tasks encompassed in our UniFashion framework: cross-modal retrieval, text-guided image retrieval, fashion image captioning, and fashion image generation. Model inputs highlighted with a light yellow background and outputs denoted by a light blue background.
        }
        \label{FashionTask}
\end{figure*}


While previous works~\citep{han2023fame, zhuge2021kaleido} in the fashion domain have also proposed using a single model for solving multiple tasks, they ignore image generation tasks. 
Besides, for fashion tasks such as try-on~\citep{choi2021viton} and fashion design~\citep{baldrati2023multimodal}, it is generally required to generate target images based on multimodal input. However, previous works~\citep{baldrati2023multimodal} in fashion image generation typically adopt the CLIP text encoder for encoding text information. This approach may not effectively capture the textual context due to the limitations of the text encoder, as noted by \citet{saharia2022photorealistic}. Hence, we posit that current studies have yet to fully explore the potential synergy between generation and retrieval.


In this work, we propose \modelname{}, which unifies retrieval and generation tasks by integrating LLMs and diffusion models, as illustrated in Figure~\ref{overview}. \modelname{} consists of three parts: The \emph{Q-Former} is crucial for amalgamating text and image input, creating multimodal learnable queries. These queries, once refined through task-specific adapters, enable the \emph{LLM} module to utilize them as soft prompts for generating captions for target images. Simultaneously, the \emph{diffusion module} utilizes the learnable queries as conditions to guide the latent diffusion model in image synthesis and editing tasks.
To enable controllable and high-fidelity generation, we propose a two-phase training strategy. In the first phase, we perform multimodal representation learning on image-text pairs datasets. We freeze Q-Former and fine-tune the LLM and diffusion modules, ensuring they develop the capability to comprehend the multimodal representations provided by Q-Former. Subsequently, in the second phase, we proceed to fine-tune \modelname{} on datasets with multimodal inputs, such as Fashion-IQ, where we freeze the LLM and diffusion modules, only tuning Q-Former. This strategy ensures that Q-Former is adept at crafting multimodal representations that effectively integrate both reference images and text inputs.


\modelname{} holds three significant advantages that address the challenges in multimodal fashion retrieval and generation:

\begin{itemize}
    \item For the first time, we conduct an in-depth study of the synergistic modeling of multimodal retrieval and generation tasks within the fashion domain,
    thoroughly exploiting the inter-task relatedness. Further, we introduce \modelname{}, a versatile, unified model that can handle all fashion tasks.
    


    \item Secondly, our model enhances performance via mutual task reinforcement. Specifically, the caption generative module aids the CIR task, while jointly training the generation and retrieval tasks improves the multimodal encoder for the diffusion module.
    
    \item Thirdly, extensive experiments on diverse fashion tasks—including cross-modal retrieval, composed image retrieval, and multimodal generation—demonstrate that our unified model significantly surpasses previous state-of-the-art methods.
\end{itemize}

\section{Preliminaries and Related Works}

\subsection{Fashion Tasks}
Fashion tasks encompass a range of image and language manipulations, including cross-modal retrieval, composed image retrieval, fashion image captioning and generation, etc. The representative tasks can be briefly divided into the following two groups.

\paragraph{Fashion Retrieval.} It generally consists of Cross-Modal Retrieval (CMR)~\citep{ma2022ei, rostamzadeh2018fashion} and composed image retrieval (CIR) tasks~\citep{baldrati2023composed, bai2023sentence}. CMR requests to efficiently retrieve the most matched image/sentence from a large candidate pool $\mathcal{D}$ given a text/image query. CIR is a special type of image retrieval with a multimodal query (a combination of a reference image and a modifying text) matched against a set of images. It retrieves a target image from a vast image database based on a reference image and a text description detailing changes to be applied to the reference image. In this scenario, a query pair $p = \{I_{R}, t\}$ is provided, where $I_{R}$ is the reference image and $t$ is the text describing the desired modifications. The challenge for this task is to accurately identify the target image $I_{T}$ that best matches the query among all potential candidates in the image corpus $\mathcal{D}$.

\paragraph{Fashion Generation.} It consists of Fashion Image Captioning (FIC) and Fashion Image Generation (FIG). {FIC}~\citep{yang2020fashion} aims to generate a descriptive caption for a product based on the visual and/or textual information provided in the input. {FIG} aims to generate images based on the multimodal input, such as try-on~\citep{choi2021viton,gou2023taming} and fashion design~\citep{baldrati2023multimodal}.


\subsection{Multimodal Language Models}
Recent research has witnessed a surge of interest in multimodal LLMs, including collaborative models~\citep{wu2023visual, yang2023mm, shen2023hugginggpt} and end-to-end methods~\citep {alayrac2022flamingo,zhao-etal-2024-easygen, li2022blip, bao2021beit, wang2022image, wang2022ofa, wang2022ofa}. More recently, some works also explore training LLMs with parameter-efficient tuning~\citep{li2023blip, zhang2023llama} and instruction tuning~\citep{instructblip, liu2023visual, ye2023mplug, zhu2023minigpt, li2023otter}. They only focus on generation tasks, while our model \modelname{} is designed as a unified framework that enables both retrieval and generation tasks.

\subsection{Diffusion Models} 
Diffusion generative models~\citep{rombach2022high,ramesh2021zero,nichol2022glide,ruiz2023dreambooth} have achieved strong results in text conditioned image generation works. Among contemporary works that aim to condition pretrained latent diffusion models, ControlNet~\citep{zhang2023adding} proposes to extend the Stable Diffusion model with an additional trainable copy part for conditioning input. In this work, we focus on the fashion domain and propose a unified framework that can leverage latent diffusion models that directly exploit the conditioning of textual sentences and other modalities such as human body poses and garment sketches. 


\subsection{Problem Formulation}

Existing fashion image retrieval and generation methods are typically designed for specific tasks, which inherently restricts their applicability to the various task forms and input/output forms in the fashion domain. To train a unified model that can handle multiple fashion tasks, our approach introduces a versatile framework capable of handling multiple fashion tasks by aligning the multimodal representation into the LLM and the diffusion model. This innovative strategy enhances the model's adaptability, and it can be represented as:
\begin{equation}
    I_{\mathrm{out}},T_{\mathrm{out}} = \mathcal{F}_{\mathcal{T}_{\mathrm{Ret}}, \mathcal{T}_{\mathrm{Gen}}}(I_{\mathrm{in}}, T_{\mathrm{in}}; \Theta),
    \label{single_task}
\end{equation}
where $\mathcal{F}_{\mathcal{T}}$ represents the unified model parameterized by $\Theta$, it consists of retrieval module ${\mathcal{T}_{Ret}}$ and generative module ${\mathcal{T}_{Gen}}$.

\begin{figure*}[t]
	\centering
	\includegraphics[width=1.0\textwidth]{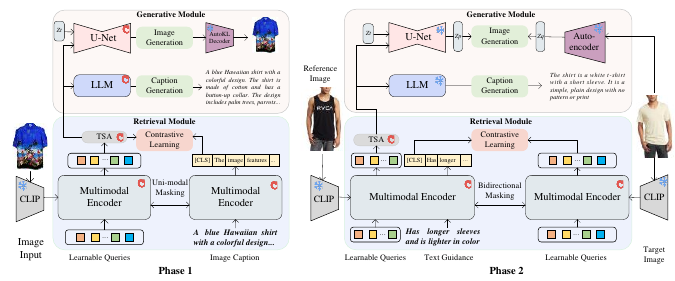}
        \caption{Overview of the training framework of our \modelname{} model. \textbf{Phase 1} - Cross-modal Pre-training: \modelname{} acquires robust cross-modal fashion representation capabilities through pre-training, leveraging both the language model and the diffusion model. \textbf{Phase 2} - Composed Multimodal Fine-tuning: The model undergoes fine-tuning to process both image and text inputs, refining its ability to learn composed modal representations. This is achieved by aligning the multimodal encoder with the LLM and the diffusion model for enhanced performance.
        }
        \label{overview}
\end{figure*}

\section{Proposed Model: \modelname{}
}

In this section, we introduce the \modelname{} to unify the fashion retrieval and generation tasks into a single model. 
By combining \textbf{retrieval and generative modules},
the proposed \modelname{} employs a \textbf{two-stage} training strategy to capture relatedness between image and language information. 
Consequently, it can seamlessly switch between two operational modes for cross-modal tasks and composed modal tasks. 

\subsection{Phase 1: Cross-modal Pre-training}
\label{cross_modal}

In the first stage, we conduct pre-training on the retrieval and generative modules to equip the Large Language Model (LLM) and diffusion model with strong cross-modal fashion representation capabilities for the next phase. 

\subsubsection{Cross-modal Retrieval}
\label{retrieval}

For cross-modal retrieval tasks, given a batch of image caption pairs $p = \{I, C\}$, we first calculate their unimodal representations using an independent method. In particular, we adopt a lightweight Querying Transformer, i.e., Q-Former in BLIP-2~\citep{li2023blip}, to encode the multimodal inputs, as it is effective in bridging the modality gap. To avoid information leaks, we employ a unimodal self-attention mask~\citep{li2023blip}, where the queries and text are not allowed to see each
other:
\begin{equation}
	\begin{aligned}
            Z_{I} &= \text{Q-Former}(I, q),\\ 
            Z_{C} &= \text{Q-Former}(C).
	\end{aligned}
 \label{eq:qformer}
\end{equation}
where the output sequence $Z_{I}$ is the encoding result of an initialized learnable query $q$ with the input image and $Z_{C}$ is the encoded caption, which contains the embedding of the output of the [CLS] token $e_{cls}$, which is a representation of the input caption text. Since $Z_{I}$ contains multiple output embeddings (one from each query), we first compute the pairwise similarity between each query output and $e_{cls}$, and then select the highest one as the image-text similarity. In our experiments, we employ 32 queries in $q$, with each query having a dimension of 768, which is the same as the hidden dimension of the Q-Former. For cross-modal learning objective, we leverage the Image-Text Contrastive Learning (ITC) and Image-Text Matching (ITM) method. The first loss term is image-text contrastive loss, which has been widely adopted in existing text-to-image retrieval models. Specifically, the image-text contrastive loss is defined as:
\begin{equation}
\begin{small}
	\begin{aligned}
            \mathcal{L}_{\mathrm{ITC}}(X,Y) = -\frac{1}{B} \sum_{i=1}^{B} \mathrm{log} \frac{\exp[ \lambda (X_{i}^{T} \cdot Y^{i})] }{\sum_{j=1}^{B} \mathrm{exp} [\lambda ( X_{i}^{T} \cdot Y^{j})]},
	\end{aligned}
 \label{eq:itcloss}
 \end{small}
\end{equation}
where $\lambda$ is a learnable temperature parameter. ITM aims to learn fine-grained
alignment between image and text representation. It is a binary classification task where the model is asked to predict whether an image-text pair is positive (matched) or negative (unmatched), it is defined as,
\begin{equation}
\begin{small}
	\begin{aligned}
            \mathcal{L}_{\mathrm{ITM}}(X,Y) = -\frac{1}{B} \sum_{i=1}^{B} \mathrm{log} \frac{\mathrm{exp} f_{\theta}(X_{i}, Y_{i})}{\sum_{j=1}^{B} \mathrm{exp}f_{\theta}(X_{j}, Y_{i})},
	\end{aligned}
 \label{eq:itmloss}
 \end{small}
\end{equation}
Then, we maximize their similarities via symmetrical contrastive loss:
\begin{equation}
	\begin{aligned}
            \mathcal{L}_{\mathrm{cross}} = \mathcal{L}_{\mathrm{ITC}}(t_{c}, Z_{I}) + \mathcal{L}_{\mathrm{ITM}}(Z_{C}, Z_{I}),
	\end{aligned}
 \label{eq:crossloss}
\end{equation}

\subsubsection{Cross-modal Generation}
\label{CMG}

As depicted in Fig.~\ref{overview}, after the learnable queries $q$ pass through the multimodal encoder, they are capable of integrating the visual information with textual guidance. However, in Section~\ref{retrieval}, we did not specify a learning target for $q$. Empirically, the $q$ that has been merged with the reference image and edited text information should be equivalent to the encoding of the target image. This implies that we should be able to reconstruct the target image and its caption based on $q$. In this section, we will employ generative objectives to improve the representation of augmented $q$.

In the first stage, we connect the Q-Former (equipped with a frozen image encoder) to a Large Language Model (LLM) to harness the LLM's prowess in language generation, and to a diffusion model to exploit its image generation capabilities. Notably, we exclusively train the model using image-text pairs throughout this process. As depicted in Figure~\ref{overview}, we employ a Task Specific Adapter (TSA) layer to linearly project the output query embeddings $q$ to match the dimensionality of the embeddings used by the LLM and diffusion model. In this stage, we freeze the parameters of the Q-Former and fine-tune only the adapter layers, connecting LLM and diffusion models. This approach allows us to develop a discriminative model that can evaluate whether queries $q$ can generate the target image and its corresponding caption.

\textbf{Target Caption Generation.}
The adapter layer is placed before the LLM to map the output of Q-Former to the text embedding space of the LLM. To synchronize the space of Q-Former with that of the LLM, we propose to use the image-grounded text generation (ITG) objective to drive the model to generate texts based on the input image by computing the auto-regressive loss:
\begin{equation}
	\begin{aligned}
		\mathcal{L}_\text{ITG} = -\frac{1}{L} \sum_{l=1}^{L}\log p_{\phi}(w_{l}^g|w_{<l}^{g},f_{\theta}(q)),
	\end{aligned}
        \label{llmloss}
\end{equation}
where $w^{g}=(w_1^g,...,w_L^g)$ represents the ground-truth caption of image $I$ with length $L$, $q=\text{Q-Former}(I,q)$, $\phi$ denotes the LLM's parameters, and $\theta$ denotes the text adapter layers' parameters.

\textbf{Target Image Generation.}
In the first stage, our task also aims to reconstruct the image $\hat{I_{T}}$ from $q$. As in standard latent diffusion models, given an encoded input $\mathbf{x}$, the proposed denoising network is trained to predict the noise stochastically added to $\mathbf{x}$. The corresponding objective function can be specified as:
\begin{equation}
	\begin{aligned}
		\mathcal{L}_{\mathrm{q2I}} &= 
  \mathbb{E}_{\bm{\epsilon}^{y},\mathbf{x}_{0}}[
  \Vert
  \bm{\epsilon}^{x} - \bm{\epsilon}_{\eta}^{x}(\mathbf{x}_{t^{x}},f_{\zeta}(q),t^{x})\Vert^{2}],
	\end{aligned}
 \label{eq:q2i}
\end{equation}
where $\eta$ denotes the u-net models' parameters and $\zeta$ denotes the image adapter layers' parameters. The overall loss in the first stage can be expressed:
\begin{equation}
    \mathcal{L}_{\mathrm{ph1}} = \mathcal{L}_{\mathrm{cross}} + \mathcal{L}_{\mathrm{ITG}}+\mathcal{L}_{\mathrm{q2T}}.
 \label{eq:stage1}
\end{equation}
After the first training stage, we can leverage the LLM and diffusion model as discriminators to guide the generation of composed queries. 


\subsection{Phase 2: Composed Multimodal Fine-tuning}
In this phase, the inputs are reference image and guidance text, and we fine-tune the model for composed multimodal retrieval and generation tasks.
\label{composed_modal}
\subsubsection{Composed Image Retrieval
}
For CIR task, the target image $I_{T}$ generally encompasses the removal of objects and the modification of attributes in the reference image. To solve this problem, as depicted in Fig.~\ref{overview}, the multimodal encoder is utilized to extract features from the reference image and the guide text. It joint embeds the given pair $p = \{I_{R}, t\}$ in a sequential output. Specifically, a set of learnable queries $q$ concatenated with text guidance $t$ is introduced to interact with the features of the reference image. Finally, the output of Q-Former is the multimodal synthetic prompt $Z_{R}$. We use a bi-directional self-attention mask, similar to the one used in BLIP2~\citep{li2023blip}, where all queries and texts can attend to each other. The output query embeddings $Z_{R}$ thus capture multimodal information:
\begin{equation}
	\begin{aligned}
            Z_{R} &= \text{Q-Former}(I_{R}, t, q_{R}),\\ 
            Z_{T} &= \text{Q-Former}(I_{T}, q_{T}).
	\end{aligned}
 \label{eq:multiqformer}
\end{equation}
Noting that the output sequence $Z_{R}$ consists of learnable queries $q$ and encoded text guidance $\bm{t}$, which includes $e_{cls}$, the embedding of the output of the [CLS] token. On the other hand, the target image's output sequence $Z_{T}$ consists only of learnable queries. Therefore, we can use $Z_{R}$ as a representation that incorporates information from the reference image and the guidance text and align it with the features of the target image $Z_{T}$. Moreover, as \modelname{} acquires the ability to generate captions for images from Sec.~\ref{CMG}, we can generate captions for the candidate images and use $e_{cls}$ to retrieve the caption $Z_{C}$ of the target image.
Then, the final contrastive loss for the CIR task is:
\begin{equation}
	\begin{aligned}
            \mathcal{L}_{\mathrm{cir}} = \mathcal{L}_{\mathrm{ITC}}(e_{cls}, Z_{T}) &+ \mathcal{L}_{\mathrm{ITC}}(e_{cls}, Z_{C}) \\ &+\mathcal{L}_{\mathrm{ITM}}(\bm{t}, Z_{T}),
	\end{aligned}
 \label{eq:cirloss}
\end{equation}

\subsubsection{Composed Multimodal Generation
}
\label{multimodel design}

For these generation tasks, we freeze the LLM parameters and tune the parameters of the task-specific adapters, the diffusion model, and the Q-Former. The loss function for the target image's caption generation is formulated in a way that is similar to Eq.~\ref{llmloss}:
\begin{equation}
	\begin{aligned}
		\mathcal{L}_\text{ITG} = -\frac{1}{L} \sum_{l=1}^{L}\log p_{\phi}(w_{l}^g|w_{<l}^{g},f_{\theta}(q_R)),
	\end{aligned}
        \label{llmloss2}
\end{equation}
The loss function for the target image generation is formulated in a way that is similar to Eq.~\ref{eq:q2i}:
\begin{equation}
	\begin{aligned}
		\mathcal{L}_{\mathrm{q2I}} &= 
  \mathbb{E}_{\bm{\epsilon}^{y},\mathbf{x}_{0}}[
  \Vert
  \bm{\epsilon}^{x} - \bm{\epsilon}_{\eta}^{x}(\mathbf{x}_{t^{x}},f_{\zeta}(q_{R}),t^{x})\Vert^{2}],
	\end{aligned}
 \label{eq:q2i2}
\end{equation}
The overall loss in the second stage can be expressed as:
\begin{equation}
    \mathcal{L}_{\mathrm{stage2}} = \mathcal{L}_{\mathrm{cir}}+\mathcal{L}_{\mathrm{ITG}}+\mathcal{L}_{\mathrm{q2I}}.
 \label{eq:stage2}
\end{equation}

\subsection{Instruction-Tuning LLMs for Different Caption Style}
\label{finetuning llm with instructions}

\citeauthor{liu2023visual}'s work shows that LLMs have the potential to handle multimodal tasks based on text description of images. Due to the different styles of captions in different fashion datasets, we adopt different instructions to tune the LLM so that it can generate captions of different styles.

We designed different instructions for different datasets and tasks, as shown in Table~\ref{instruction templates}. General instruction template is denoted as follows:

\noindent {USER: <Img><queries></Img> + Instruction. Assistant: <answer>.}

For the <image> placeholder, we substitute it with the output of Multimodal Encoder. 
To avoid overfitting to the specific task and counteract the model's inclination to generate excessively short outputs, we have devised specific instructions, which enable the LLM to produce concise responses when necessary.

\begin{table*}
	\centering
	\fontsize{8}{8}\selectfont
	\begin{tabular}{l|ccccccc}
		\toprule
		\multirow{2}{*}{\bf Model}&
		\multicolumn{3}{c}{\bf Image to Text}&
		\multicolumn{3}{c}{\bf Text to Image}&\multirow{2}{*}{\bf Mean}\cr
		\cmidrule(lr){2-4}\cmidrule(lr){5-7}
		&R@1&R@5&R@10&R@1&R@5&R@10&\cr
		\midrule
		{ FashionBERT~\citep{li2022blip} }&23.96&46.31&52.12&26.75&46.48&55.74&41.89\cr
        { OSCAR~\citep{alayrac2022flamingo} }&23.39&44.67&52.55&25.10&49.14&56.68&41.92\cr
		{ KaledioBERT~\citep{li2023blip}}&27.99&60.09&68.37&33.88&60.60&68.59&53.25\cr
		{ EI-CLIP~\citep{li2023blip}}&38.70&72.20&84.25&40.06&71.99&82.90&65.02\cr
            { MVLT~\citep{instructblip}}&33.10&77.20&91.10&34.60&78.00&89.50&67.25\cr
             { FashionViL~\citep{zhu2023minigpt}}&65.54&91.34&96.30&61.88&87.32&93.22&82.60\cr
             { FAME-ViL~\citep{liu2023visual}}&65.94&91.92&97.22&62.86&87.38&93.52&83.14\cr
		\cmidrule(lr){1-8}
		{\bf \modelname{} (Ours)}&{\bf 71.44}&{\bf 93.79}&{\bf 97.51}&{\bf 71.41}&{\bf 93.69}&{\bf 97.47}&{\bf 87.55}\cr
		\bottomrule
	\end{tabular}
 	\caption{Performance comparison of \modelname{} and baseline models on the FashionGen dataset for cross-modal retrieval tasks. }
        \label{tab:performance_comparison}
\end{table*}

\begin{table}[t]
	\centering
	\fontsize{7.2}{8}\selectfont
	\begin{tabular}{l|cccccccccc}
		\toprule
		\multirow{2}{*}{\bf Model}
  &
		\multicolumn{4}{c}{\bf Image Captioning}\cr
		\cmidrule(lr){2-5}
		&BLEU-4&METEOR&ROUGE-L&CIDEr\cr
		\midrule
		{FashionBERT}&3.30&9.80&29.70&30.10\cr
		OSCAR&4.50&10.90&30.10&30.70\cr
            KaleidoBERT&5.70&12.80&32.90&32.60\cr
            FashionViL&16.18&25.60&37.23&39.30\cr
            FAME-ViL&30.73&25.04&{\bf55.83}&150.4\cr
        \specialrule{0.05em}{0.3em}{0.3em}
		{\bf \modelname{}}&{\bf 35.53}&{\bf 29.32}&{54.59}&{\bf 169.5}\cr
	\bottomrule
	\end{tabular}
        \caption{The Performance of \modelname{} in the image captioning task on the FashionGen dataset.}
        \label{tab:image_captioning}
\end{table}

\begin{table*}[t]
	\centering
	\fontsize{7.2}{8}\selectfont
	\begin{tabular}{l|cccccccccc}
		\toprule
		\multirow{2}{*}{\bf Model} &
		\multicolumn{4}{c}{\bf Modalities}&\multicolumn{3}{c}{\bf Metrics}\cr
		\cmidrule(lr){2-5} \cmidrule(lr){6-8}
		&Text&Sketch&Pose&Cloth& FID$ \downarrow$ & KID $\downarrow$ & CLIP-S\cr
		\midrule
            \emph{try-on task} \cr
		VITON-HD~\citep{choi2021viton}&\usym{2717}&\usym{2717}&\checkmark&\checkmark&12.12&3.23&-\cr
		Paint-by-Example~\citep{yang2023paint}&\usym{2717}&\usym{2717}&\checkmark&\checkmark&11.94&3.85&-\cr
            GP-VTON~\citep{xie2023gp}&\usym{2717}&\usym{2717}&\checkmark&\checkmark&13.07&4.66&-\cr
            StableVITON~\citep{kim2024stableviton}&\usym{2717}&\usym{2717}&\checkmark&\checkmark&\bf 8.23&\bf 0.49&-\cr
            {\bf \modelname{}} (Ours)&\usym{2717}&\usym{2717}&\checkmark&\checkmark&\underline{8.42}&\underline{0.67}&-\cr
        \specialrule{0.05em}{0.3em}{0.3em}
            \emph{fashion design task}\cr
            SDEdit~\citep{meng2021sdedit}&\checkmark&\checkmark&\checkmark&\usym{2717}&15.12&5.67&28.61\cr
            MGD~\citep{baldrati2023multimodal}&\checkmark&\checkmark&\checkmark&\usym{2717}&\underline{12.81}&\underline{3.86}&\underline{30.75}\cr
		{\bf \modelname{}} (Ours)&\checkmark&\checkmark&\checkmark&\usym{2717}&\bf 12.43&\bf 3.74&\bf 31.29\cr
	\bottomrule
	\end{tabular}
        \caption{Performance analysis of unpaired settings on the VITON-HD and MGD datasets across different input modalities.}
        \label{tab:performance_comparison_viton}
\end{table*}

\begin{table*}
	\centering
	\fontsize{8.5}{8}\selectfont
	\begin{tabular}{l|cc|cc|cc|cc|c}
		\toprule
  
        \multirow{2}{*}{\bf Model}&
        \multicolumn{2}{c}{\bf Dress}&
        \multicolumn{2}{c}{\bf Shirt}&\multicolumn{2}{c}{\bf Toptee}&\multicolumn{3}{c}{\bf Average}\cr
        
        \cmidrule(lr){2-3}\cmidrule(lr){4-5}\cmidrule(lr){6-7}\cmidrule(lr){8-10}
        &R@10&R@50&R@10&R@50&R@10&R@50&R@10&R@50&Avg.\cr
        \midrule
         FashionVLP~\citep{goenka2022fashionvlp} &32.42&60.29&31.89&58.44&38.51&68.79&34.27&62.51&48.39\cr
         CASE~\citep{levy2023data} &47.44&69.36&48.48&70.23&50.18&72.24&48.79&70.68&59.74\cr
         AMC~\citep{zhu2023amc} &31.73&59.25&30.67&59.08&36.21&66.06&32.87&61.64&47.25\cr
         CoVR-BLIP~\citep{ventura2024covr}&44.55&69.03&48.43&67.42&52.60&74.31&48.53&70.25&59.39\cr
         
         MGUR~\citep{chen2022composed}&32.61&61.34&33.23&62.55&41.40&72.51&35.75&65.47&50.61\cr
         LinCIR~\citep{gu2024language}&38.08&60.88&46.76&65.11&50.48&71.09&45.11&65.69&55.4\cr
         CMAP~\citep{li2024cross}&36.44&64.25&34.83&60.06&41.79&69.12&37.64&64.42&51.03\cr
         CLIP4CIR~\citep{baldrati2023composed}&33.81&59.40&39.99&60.45&41.41&65.37&38.32&61.74&50.03\cr
         FAME-ViL~\citep{han2023fame}&42.19&67.38&47.64&68.79&50.69&73.07&46.84&69.75&58.29\cr
         TG-CIR~\citep{wen2023target}&45.22&69.66&52.60&72.52&56.14&77.10&51.32&73.09&58.05\cr
         Re-ranking~\citep{liu2023candidate}&48.14&71.43&50.15&71.25&55.23&76.80&51.17&73.13&62.15\cr
         SPRC ~\citep{bai2023sentence}&49.18&\underline{72.43}&55.64&73.89&\underline{59.35}&\underline{78.58}&54.92&\underline{74.97}&64.85\cr
        \rowcolor{mygray} \modelname{} w/o cap&\underline{49.65}&72.17&\underline{56.88}&\underline{74.12}&59.29&78.11&\underline{55.2}7&74.80&\underline{65.04}\cr
        \rowcolor{mygray} \modelname{} w/o img&32.49&49.11&44.70&59.63&43.16&60.26&40.12&56.33&48.22\cr
        \rowcolor{mygray} \modelname{} &\bf{53.72}&\bf 73.66&\bf 61.25&\bf 76.67&\bf 61.84&\bf 80.46&\bf 58.93&\bf 76.93&\bf 67.93\cr
		\bottomrule
	\end{tabular}
 
        \caption{Comparative evaluation of \modelname{} and variants and baseline models on the Fashion-IQ dataset for composed image retrieval task. Best and second-best results are highlighted in bold and underlined, respectively.
        }
        \label{CIR results}
\end{table*}

\begin{table}[t]
	\centering
	\fontsize{8.5}{8}\selectfont
	\begin{tabular}{l|c|c|c|c}
		\toprule
  
        {\bf Model}&
        {\bf CMR}&
        {\bf CIR}&{\bf FIC}&{\bf FIG}\cr
        
        \midrule
        
        Base&87.38&64.76&-&-\cr
        Base+LLM &87.49&65.04&\bf 36.21&-\cr
        Base+LLM w/ cap &87.49&66.83&\bf 36.21&-\cr
        Base+LLM+diff. &87.55&\bf 67.93&35.53&12.43\cr
		\bottomrule
	\end{tabular}
        \caption{Ablation study and analysis of \modelname{} across FashionGen, Fashion-IQ, and VITON-HD Datasets. Metrics reported include average image-to-text and text-to-image recall for cross-modal retrieval (CMR), average recall for composed image retrieval (CIR), BLEU-4 for Fashion Image Captioning, and FID for Fashion image generation (FIG). 
        }
        \label{ablation detail}
\end{table}

\section{Experiments}

\subsection{Experimental Setup}
We initialize the multimodal encoder using BLIP2's Q-Former. Following the approach of LLaVA-1.5~\citep{liu2023visual}, we initialize the LLM from Vicuna-1.5~\citep{zheng2023judging}. For the diffusion module, we adopt the autoencoder and denoising U-Net from Stable Diffusion v1.4, as utilized in StableVITON. The weights of the U-Net are initialized from Paint-by-Example. To achieve more refined person textures, we employ a VAE that has been fine-tuned on the VITONHD dataset, as done in StableVITON. The statistics of the two-stage datasets can be found in Table~\ref{VQA datasets}. For cross-modal retrieval, we evaluated \modelname{} on FashionGen validation set. For the image captioning task, \modelname{} is evaluated in the FashionGen dataset. For the composed image retrieval task, we evaluated the Fashion-IQ validation set. To maintain consistency with previous work, for the composed image generation task, we fine-tuned \modelname{} and evaluated it on the VITON-HD and MGD datasets. More details can be found in Appendix~\ref{implementation}.

\paragraph{Phase 1:} For multimodal representation learning, we follow BLIP2 and pretrain the Q-Former on fashion image-text pairs. To adapt the model for multimodal generation, we freeze the parameters of Q-Former and fine-tune the MLLM and diffusion model with their task specific adapters separately. Due to the different styles of captions in different fashion datasets, we adopt the approach of instruction tuning to train the LLM so that it can generate captions of different styles. More details can be found in Appendix~\ref{finetuning llm with instructions}.

\paragraph{Phase 2:} In order to make \modelname{} have the composed retrieval and generation abilities, we freeze the parameters of LLM and diffusion model, only fine-tune the multimodal encoder.

\subsection{Datasets}
We test the effectiveness of \modelname{} by experimenting on different tasks including fashion image captioning, cross-modal retrieval, composed image retrieval and composed image generation.

We use the FashionGen and FshaionIQ~\citep{lin2014microsoft} datasets for retrieval tasks. FashionGen contains 68k fashion products accompanied by text descriptions. Each product includes
1 - 6 images from different angles, resulting in 260.5k image-text pairs for training and 35.5k for testing. Fashion-IQ contains 18k training triplets (that is, reference image, modifying text, target image) and 6k validation triplets over three categories: Dress, Shirt, and Toptee. Each pair (reference image, target image) is manually annotated with two modifying texts, which are concatenated.

For fashion image captioning tasks, we utilize the FashionGen~\citep{zang2021photochat} dataset. Additionally, to enhance our model's capability in the CIR task, which involves the ability to retrieve captions for target images, we have annotated images from the training set of Fashion-IQ. Recognizing that manually annotating all the images would be time-consuming and resource-intensive, we draw inspiration from the success of recent MLLM models such as LLaVA in text-annotation tasks, and propose leveraging LLaVA 1.5 (13B) to semi-automatically annotate the dataset. More details can be found in Appendix~\ref{dataset detail}.

\subsection{Evaluation Methods}
We compare our models with previous state-of-the-art methods on each task. For extensive and fair comparisons, all prior competitors are based on large-scale pre-trained models.

\paragraph{Cross-modal Retrieval Evaluation.} We consider both image-to-text retrieval and text-to-image retrieval with random 100 protocols used by previous methods. 100 candidates are randomly sampled from the same category to construct a retrieval database. The goal is to locate the positive match depicting the same garment instance from these 100 same-category negative matches. We utilize Recall@K as the evaluation metric, which reflects the percentage of queries whose true target ranked within the top K candidates.

\paragraph{Fashion Image Captioning Evaluation.} For evaluating the performance of caption generation, we utilize BLEU-4, METEOR, ROUGE-L, and CIDEr as metrics.

\paragraph{Composed Fashion Image Retrieval Evaluation.} We compare our \modelname{} with CIR methods and the FAME-ViL model of V + L that is oriented towards fashion in the original protocol used by Fashion-IQ. For this task, we also utilize Recall@K as the evaluation metric.

\paragraph{Composed Fashion Image Generation Evaluation.} We compare our \modelname{} with try-on methods on VITON-HD dataset and fashion design works on MGD dataset. To evaluate the quality of image generation, we use the Frechet Inception Distance (FID) score to measure the divergence between two multivariate normal distributions and employ the CLIP Score (CLIP-S) provided in the TorchMetrics library to assess the adherence of the image to the textual conditioning input (for fashion design task).

\subsection{Comparative Analysis of Baselines and Our Method}
\textbf{\modelname{} exhibits superior performance across all datasets compared to baselines.}
Tab.~\ref{tab:performance_comparison} presents the evaluation results for each baseline and our models in FashionGen data sets for cross-modal retrieval. \modelname{} outperforms most of the baseline models on both the text-to-image and image-to-text tasks. Following FAME-ViL, we also adopt a more challenging and practical protocol that conducts retrieval on the entire product set, which is in line with actual product retrieval scenarios. In Tab.~\ref{tab:image_captioning}, we performed a comparison between our \modelname{} and other baselines on the FashionGen dataset for the image captioning task. By integrating the powerful generative ability of the LLM, our model performed significantly better than the traditional multimodal models in this task. In Tab.~\ref{CIR results}, we conducted a comparison between our \modelname{} and CIR-specialist methods. Our findings are in line with those of Tab.~\ref{tab:performance_comparison}.

\textbf{After fine-tuning \modelname{} on image generation/editing tasks with multimodal inputs, it exhibits outstanding performance.}
Tab.~\ref{tab:performance_comparison_viton} evaluates the quality of the generated image of \modelname{} in the VITON-HD unpaired setting. In order to verify that our model can achieve good results in a variety of modal inputs, we have conducted tests, respectively, on the traditional try-on task and the fashion design task proposed in MGD. For a fair evaluation with baselines, all the models are trained at a 512 × 384 resolution. To confirm the efficacy of our approach, we assess the realism using FID and KID score on all the tasks and using CLIP-S score for fashion design task. As can be seen, the proposed \modelname{} model consistently outperforms competitors in terms of realism (i.e., FID and KID) and coherence with input modalities (i.e., CLIP-S), indicating that our method can better encode multimodal information. Meanwhile, although our model is slightly lower than StableVITON on the try-on task, this is because we froze the parameters of the diffusion model on the try-on task and only fine-tuned the Q-former part, but it can still achieve top2 results. The visual results can be found in Appendix~\ref{fashion_design}.

\subsection{Ablation Study}
\label{ablation study}

\textbf{\modelname{} allows for more flexible execution of multimodal composed tasks.} In Tab.~\ref{CIR results}, we also carry out ablation studies on different retrieval methods. Since \modelname{} is capable of generating captions, for the CIR task, we initially utilize \modelname{} to generate the captions of candidate images and then conduct the image retrieval task (denoted as \modelname{} w/o cap) and the caption retrieval task (denoted as \modelname{} w/o img). We find that our single-task variant has already achieved superior performance in the relevant field. Furthermore, due to the generative ability of our model, the pregenerated candidate library optimizes the model's performance in this task. For specific implementation details, please refer to Appendix~\ref{dataset detail}.

\textbf{We investigate the impact of different modules in \modelname{} on various fashion tasks.}
In Tab.~\ref{ablation detail}, we perform an ablation study on the proposed model architecture, with a focus on LLM and diffusion models. For comparison on the cross-modal retrieval task (CMR), we design the base model as directly fine-tuning BLIP2 without any new modules. The results indicate that the base model performs relatively well on this task and that the introduction of other modules does not lead to significant improvements. However, in the CIR task, the introduction of LLM and diffusion models as supervision can lead to significant improvements, especially when utilizing pregenerated captions by \modelname{} to assist in retrieval, resulting in greater benefits. At the same time, we note that, after introducing the diffusion model, it may have some negative impact on the model's image captioning ability, possibly due to the inherent alignment differences between LLM and the diffusion model.


\section{Conclusion}


We have introduced \modelname{}, a unified framework designed to tackle challenges in multimodal generation and retrieval within the fashion domain. By integrating embedding and generative tasks using a diffusion model and LLM, UniFashion enables controllable, high-fidelity generation, significantly outperforming previous single-task state-of-the-art models across various fashion tasks. Our model's adaptability in handling complex vision-language tasks demonstrates its potential to enhance e-commerce scenarios and fashion-related applications. This study highlights the importance of exploring the learning synergy between multimodal generation and retrieval, offering a promising direction for future research in the fashion domain.




\section*{Limitations}
In this section, we discuss limitations of our work and offer further insights into research within the fashion domain.

\textbf{Computational Requirements.}
\modelname{} integrates multiple complex modules, including Q-Former, LLM, and diffusion models, which result in higher computational complexity during training. However, during the inference stage, the computational complexity of \modelname{} is comparable to that of current state-of-the-art models. For retrieval tasks, only the Q-Former module is needed to calculate the similarity between the input image or text and the pre-stored candidate features in the database, eliminating the need to utilize the LLM and diffusion model components for inference. For composed image generation tasks, such as fashion design, our model relies on diffusion processes, which may take longer. In our experiments, we tested the performance of our model on an A100 (80G) GPU. During inference, using 1000 examples from the VITON-HD dataset, \modelname{} took approximately 3.15 seconds per image generation. We believe exploring more efficient sampling methods, such as DPM-Solver++~\cite{lu2022dpm}, could improve the overall efficiency of \modelname{}.


\section*{Acknowledgements}
We thank the anonymous reviewers for their valuable feedback. This research was partially supported by the grant of HK ITF ITS/359/21FP.

\bibliography{custom}

\begin{thebibliography}{71}
\expandafter\ifx\csname natexlab\endcsname\relax\def\natexlab#1{#1}\fi

\bibitem[{Alayrac et~al.(2022)Alayrac, Donahue, Luc, Miech, Barr, Hasson, Lenc, Mensch, Millican, Reynolds et~al.}]{alayrac2022flamingo}
Jean-Baptiste Alayrac, Jeff Donahue, Pauline Luc, Antoine Miech, Iain Barr, Yana Hasson, Karel Lenc, Arthur Mensch, Katherine Millican, Malcolm Reynolds, et~al. 2022.
\newblock Flamingo: a visual language model for few-shot learning.
\newblock \emph{Advances in Neural Information Processing Systems}, 35:23716--23736.

\bibitem[{Bai et~al.(2023)Bai, Xu, Liu, Khan, Khan, Zuo, Goh, and Feng}]{bai2023sentence}
Yang Bai, Xinxing Xu, Yong Liu, Salman Khan, Fahad Khan, Wangmeng Zuo, Rick Siow~Mong Goh, and Chun-Mei Feng. 2023.
\newblock Sentence-level prompts benefit composed image retrieval.
\newblock \emph{arXiv preprint arXiv:2310.05473}.

\bibitem[{Baldrati et~al.(2022)Baldrati, Bertini, Uricchio, and Del~Bimbo}]{baldrati2022effective}
Alberto Baldrati, Marco Bertini, Tiberio Uricchio, and Alberto Del~Bimbo. 2022.
\newblock Effective conditioned and composed image retrieval combining clip-based features.
\newblock In \emph{Proceedings of the IEEE/CVF conference on computer vision and pattern recognition}, pages 21466--21474.

\bibitem[{Baldrati et~al.(2023{\natexlab{a}})Baldrati, Bertini, Uricchio, and Del~Bimbo}]{baldrati2023composed}
Alberto Baldrati, Marco Bertini, Tiberio Uricchio, and Alberto Del~Bimbo. 2023{\natexlab{a}}.
\newblock Composed image retrieval using contrastive learning and task-oriented clip-based features.
\newblock \emph{ACM Transactions on Multimedia Computing, Communications and Applications}, 20(3):1--24.

\bibitem[{Baldrati et~al.(2023{\natexlab{b}})Baldrati, Morelli, Cartella, Cornia, Bertini, and Cucchiara}]{baldrati2023multimodal}
Alberto Baldrati, Davide Morelli, Giuseppe Cartella, Marcella Cornia, Marco Bertini, and Rita Cucchiara. 2023{\natexlab{b}}.
\newblock Multimodal garment designer: Human-centric latent diffusion models for fashion image editing.
\newblock In \emph{Proceedings of the IEEE/CVF International Conference on Computer Vision}, pages 23393--23402.

\bibitem[{Bao et~al.(2021)Bao, Dong, Piao, and Wei}]{bao2021beit}
Hangbo Bao, Li~Dong, Songhao Piao, and Furu Wei. 2021.
\newblock Beit: Bert pre-training of image transformers.
\newblock In \emph{International Conference on Learning Representations}.

\bibitem[{Brown et~al.(2020)Brown, Mann, Ryder, Subbiah, Kaplan, Dhariwal, Neelakantan, Shyam, Sastry, Askell et~al.}]{brown2020language}
Tom Brown, Benjamin Mann, Nick Ryder, Melanie Subbiah, Jared~D Kaplan, Prafulla Dhariwal, Arvind Neelakantan, Pranav Shyam, Girish Sastry, Amanda Askell, et~al. 2020.
\newblock Language models are few-shot learners.
\newblock \emph{Advances in neural information processing systems}, 33:1877--1901.

\bibitem[{Chen et~al.(2022)Chen, Zheng, Ji, Qu, and Chua}]{chen2022composed}
Yiyang Chen, Zhedong Zheng, Wei Ji, Leigang Qu, and Tat-Seng Chua. 2022.
\newblock Composed image retrieval with text feedback via multi-grained uncertainty regularization.
\newblock \emph{arXiv preprint arXiv:2211.07394}.

\bibitem[{Chiang et~al.(2023)Chiang, Li, Lin, Sheng, Wu, Zhang, Zheng, Zhuang, Zhuang, Gonzalez, Stoica, and Xing}]{vicuna2023}
Wei-Lin Chiang, Zhuohan Li, Zi~Lin, Ying Sheng, Zhanghao Wu, Hao Zhang, Lianmin Zheng, Siyuan Zhuang, Yonghao Zhuang, Joseph~E. Gonzalez, Ion Stoica, and Eric~P. Xing. 2023.
\newblock \href {https://lmsys.org/blog/2023-03-30-vicuna/} {Vicuna: An open-source chatbot impressing gpt-4 with 90\%* chatgpt quality}.

\bibitem[{Choi et~al.(2021)Choi, Park, Lee, and Choo}]{choi2021viton}
Seunghwan Choi, Sunghyun Park, Minsoo Lee, and Jaegul Choo. 2021.
\newblock Viton-hd: High-resolution virtual try-on via misalignment-aware normalization.
\newblock In \emph{Proceedings of the IEEE/CVF conference on computer vision and pattern recognition}, pages 14131--14140.

\bibitem[{Dai et~al.(2023)Dai, Li, Li, Tiong, Zhao, Wang, Li, Fung, and Hoi}]{instructblip}
Wenliang Dai, Junnan Li, Dongxu Li, Anthony Meng~Huat Tiong, Junqi Zhao, Weisheng Wang, Boyang Li, Pascale Fung, and Steven Hoi. 2023.
\newblock Instructblip: Towards general-purpose vision-language models with instruction tuning.

\bibitem[{Dong et~al.(2023)Dong, Han, Peng, Qi, Ge, Yang, Zhao, Sun, Zhou, Wei et~al.}]{dong2023dreamllm}
Runpei Dong, Chunrui Han, Yuang Peng, Zekun Qi, Zheng Ge, Jinrong Yang, Liang Zhao, Jianjian Sun, Hongyu Zhou, Haoran Wei, et~al. 2023.
\newblock Dreamllm: Synergistic multimodal comprehension and creation.
\newblock \emph{arXiv preprint arXiv:2309.11499}.

\bibitem[{Feng et~al.(2023)Feng, Lu, Liu, Zhan, and Wu}]{feng2023towards}
Yujie Feng, Zexin Lu, Bo~Liu, Liming Zhan, and Xiao-Ming Wu. 2023.
\newblock Towards llm-driven dialogue state tracking.
\newblock In \emph{Proceedings of the 2023 Conference on Empirical Methods in Natural Language Processing}, pages 739--755.

\bibitem[{Gao et~al.(2020)Gao, Jin, Chen, Qiu, Li, Wei, Hu, and Wang}]{gao2020fashionbert}
Dehong Gao, Linbo Jin, Ben Chen, Minghui Qiu, Peng Li, Yi~Wei, Yi~Hu, and Hao Wang. 2020.
\newblock Fashionbert: Text and image matching with adaptive loss for cross-modal retrieval.
\newblock In \emph{Proceedings of the 43rd International ACM SIGIR Conference on Research and Development in Information Retrieval}, pages 2251--2260.

\bibitem[{Goenka et~al.(2022)Goenka, Zheng, Jaiswal, Chada, Wu, Hedau, and Natarajan}]{goenka2022fashionvlp}
Sonam Goenka, Zhaoheng Zheng, Ayush Jaiswal, Rakesh Chada, Yue Wu, Varsha Hedau, and Pradeep Natarajan. 2022.
\newblock Fashionvlp: Vision language transformer for fashion retrieval with feedback.
\newblock In \emph{Proceedings of the IEEE/CVF Conference on Computer Vision and Pattern Recognition}, pages 14105--14115.

\bibitem[{Gou et~al.(2023)Gou, Sun, Zhang, Si, Qian, and Zhang}]{gou2023taming}
Junhong Gou, Siyu Sun, Jianfu Zhang, Jianlou Si, Chen Qian, and Liqing Zhang. 2023.
\newblock Taming the power of diffusion models for high-quality virtual try-on with appearance flow.
\newblock In \emph{Proceedings of the 31st ACM International Conference on Multimedia}, pages 7599--7607.

\bibitem[{Goyal et~al.(2017)Goyal, Khot, Summers-Stay, Batra, and Parikh}]{goyal2017making}
Yash Goyal, Tejas Khot, Douglas Summers-Stay, Dhruv Batra, and Devi Parikh. 2017.
\newblock Making the v in vqa matter: Elevating the role of image understanding in visual question answering.
\newblock In \emph{Proceedings of the IEEE conference on computer vision and pattern recognition}, pages 6904--6913.

\bibitem[{Gu et~al.(2024)Gu, Chun, Kim, Kang, and Yun}]{gu2024language}
Geonmo Gu, Sanghyuk Chun, Wonjae Kim, Yoohoon Kang, and Sangdoo Yun. 2024.
\newblock Language-only training of zero-shot composed image retrieval.
\newblock In \emph{Proceedings of the IEEE/CVF Conference on Computer Vision and Pattern Recognition}, pages 13225--13234.

\bibitem[{Han et~al.(2023)Han, Zhu, Yu, Zhang, Song, and Xiang}]{han2023fame}
Xiao Han, Xiatian Zhu, Licheng Yu, Li~Zhang, Yi-Zhe Song, and Tao Xiang. 2023.
\newblock Fame-vil: Multi-tasking vision-language model for heterogeneous fashion tasks.
\newblock In \emph{Proceedings of the IEEE/CVF Conference on Computer Vision and Pattern Recognition}, pages 2669--2680.

\bibitem[{Han et~al.(2017)Han, Wu, Huang, Zhang, Zhu, Li, Zhao, and Davis}]{han2017automatic}
Xintong Han, Zuxuan Wu, Phoenix~X Huang, Xiao Zhang, Menglong Zhu, Yuan Li, Yang Zhao, and Larry~S Davis. 2017.
\newblock Automatic spatially-aware fashion concept discovery.
\newblock In \emph{Proceedings of the IEEE international conference on computer vision}, pages 1463--1471.

\bibitem[{Ho et~al.(2020)Ho, Jain, and Abbeel}]{ho2020denoising}
Jonathan Ho, Ajay Jain, and Pieter Abbeel. 2020.
\newblock Denoising diffusion probabilistic models.
\newblock \emph{Advances in neural information processing systems}, 33:6840--6851.

\bibitem[{Kim et~al.(2024)Kim, Gu, Park, Park, and Choo}]{kim2024stableviton}
Jeongho Kim, Guojung Gu, Minho Park, Sunghyun Park, and Jaegul Choo. 2024.
\newblock Stableviton: Learning semantic correspondence with latent diffusion model for virtual try-on.
\newblock In \emph{Proceedings of the IEEE/CVF Conference on Computer Vision and Pattern Recognition}, pages 8176--8185.

\bibitem[{Krishna et~al.(2017)Krishna, Zhu, Groth, Johnson, Hata, Kravitz, Chen, Kalantidis, Li, Shamma et~al.}]{krishna2017visual}
Ranjay Krishna, Yuke Zhu, Oliver Groth, Justin Johnson, Kenji Hata, Joshua Kravitz, Stephanie Chen, Yannis Kalantidis, Li-Jia Li, David~A Shamma, et~al. 2017.
\newblock Visual genome: Connecting language and vision using crowdsourced dense image annotations.
\newblock \emph{International journal of computer vision}, 123:32--73.

\bibitem[{Levy et~al.(2023)Levy, Ben-Ari, Darshan, and Lischinski}]{levy2023data}
Matan Levy, Rami Ben-Ari, Nir Darshan, and Dani Lischinski. 2023.
\newblock Data roaming and early fusion for composed image retrieval.
\newblock \emph{arXiv preprint arXiv:2303.09429}.

\bibitem[{Li et~al.(2023{\natexlab{a}})Li, Zhang, Chen, Wang, Yang, and Liu}]{li2023otter}
Bo~Li, Yuanhan Zhang, Liangyu Chen, Jinghao Wang, Jingkang Yang, and Ziwei Liu. 2023{\natexlab{a}}.
\newblock Otter: A multi-modal model with in-context instruction tuning.
\newblock \emph{arXiv preprint arXiv:2305.03726}.

\bibitem[{Li et~al.(2023{\natexlab{b}})Li, Li, Savarese, and Hoi}]{li2023blip}
Junnan Li, Dongxu Li, Silvio Savarese, and Steven Hoi. 2023{\natexlab{b}}.
\newblock Blip-2: Bootstrapping language-image pre-training with frozen image encoders and large language models.
\newblock \emph{arXiv preprint arXiv:2301.12597}.

\bibitem[{Li et~al.(2022)Li, Li, Xiong, and Hoi}]{li2022blip}
Junnan Li, Dongxu Li, Caiming Xiong, and Steven Hoi. 2022.
\newblock Blip: Bootstrapping language-image pre-training for unified vision-language understanding and generation.
\newblock In \emph{International Conference on Machine Learning}, pages 12888--12900. PMLR.

\bibitem[{Li et~al.(2024)Li, Xu, Jiang, Shen, Sun, and Cichocki}]{li2024cross}
Shenshen Li, Xing Xu, Xun Jiang, Fumin Shen, Zhe Sun, and Andrzej Cichocki. 2024.
\newblock Cross-modal attention preservation with self-contrastive learning for composed query-based image retrieval.
\newblock \emph{ACM Transactions on Multimedia Computing, Communications and Applications}, 20(6):1--22.

\bibitem[{Lin et~al.(2014)Lin, Maire, Belongie, Hays, Perona, Ramanan, Doll{\'a}r, and Zitnick}]{lin2014microsoft}
Tsung-Yi Lin, Michael Maire, Serge Belongie, James Hays, Pietro Perona, Deva Ramanan, Piotr Doll{\'a}r, and C~Lawrence Zitnick. 2014.
\newblock Microsoft coco: Common objects in context.
\newblock In \emph{Computer Vision--ECCV 2014: 13th European Conference, Zurich, Switzerland, September 6-12, 2014, Proceedings, Part V 13}, pages 740--755. Springer.

\bibitem[{Liu et~al.(2023{\natexlab{a}})Liu, Li, Wu, and Lee}]{liu2023visual}
Haotian Liu, Chunyuan Li, Qingyang Wu, and Yong~Jae Lee. 2023{\natexlab{a}}.
\newblock Visual instruction tuning.
\newblock \emph{arXiv preprint arXiv:2304.08485}.

\bibitem[{Liu et~al.(2024{\natexlab{a}})Liu, Dong, Xiao, Chen, Hu, Zhu, Zhu, Sakai, and Wu}]{liu2024vector}
Qijiong Liu, Xiaoyu Dong, Jiaren Xiao, Nuo Chen, Hengchang Hu, Jieming Zhu, Chenxu Zhu, Tetsuya Sakai, and Xiao-Ming Wu. 2024{\natexlab{a}}.
\newblock Vector quantization for recommender systems: A review and outlook.
\newblock \emph{arXiv preprint arXiv:2405.03110}.

\bibitem[{Liu et~al.(2022)Liu, Zhu, Dai, and Wu}]{liu2022boosting}
Qijiong Liu, Jieming Zhu, Quanyu Dai, and Xiao-Ming Wu. 2022.
\newblock Boosting deep ctr prediction with a plug-and-play pre-trainer for news recommendation.
\newblock In \emph{Proceedings of the 29th International Conference on Computational Linguistics}, pages 2823--2833.

\bibitem[{Liu et~al.(2024{\natexlab{b}})Liu, Zhu, Yang, Dai, Du, Wu, Zhao, Zhang, and Dong}]{liu2024multimodal}
Qijiong Liu, Jieming Zhu, Yanting Yang, Quanyu Dai, Zhaocheng Du, Xiao-Ming Wu, Zhou Zhao, Rui Zhang, and Zhenhua Dong. 2024{\natexlab{b}}.
\newblock Multimodal pretraining, adaptation, and generation for recommendation: A survey.
\newblock In \emph{Proceedings of the 30th ACM SIGKDD Conference on Knowledge Discovery and Data Mining}, pages 6566--6576.

\bibitem[{Liu et~al.(2021)Liu, Rodriguez-Opazo, Teney, and Gould}]{liu2021image}
Zheyuan Liu, Cristian Rodriguez-Opazo, Damien Teney, and Stephen Gould. 2021.
\newblock Image retrieval on real-life images with pre-trained vision-and-language models. in 2021 ieee.
\newblock In \emph{CVF International Conference on Computer Vision (ICCV)(2021)}, pages 2105--2114.

\bibitem[{Liu et~al.(2023{\natexlab{b}})Liu, Sun, Teney, and Gould}]{liu2023candidate}
Zheyuan Liu, Weixuan Sun, Damien Teney, and Stephen Gould. 2023{\natexlab{b}}.
\newblock Candidate set re-ranking for composed image retrieval with dual multi-modal encoder.
\newblock \emph{arXiv preprint arXiv:2305.16304}.

\bibitem[{Lu et~al.(2022)Lu, Zhou, Bao, Chen, Li, and Zhu}]{lu2022dpm}
Cheng Lu, Yuhao Zhou, Fan Bao, Jianfei Chen, Chongxuan Li, and Jun Zhu. 2022.
\newblock Dpm-solver++: Fast solver for guided sampling of diffusion probabilistic models.
\newblock \emph{arXiv preprint arXiv:2211.01095}.

\bibitem[{Ma et~al.(2022)Ma, Zhao, Lin, Kale, Wang, Yu, Gu, Choudhary, and Xie}]{ma2022ei}
Haoyu Ma, Handong Zhao, Zhe Lin, Ajinkya Kale, Zhangyang Wang, Tong Yu, Jiuxiang Gu, Sunav Choudhary, and Xiaohui Xie. 2022.
\newblock Ei-clip: Entity-aware interventional contrastive learning for e-commerce cross-modal retrieval.
\newblock In \emph{Proceedings of the IEEE/CVF Conference on Computer Vision and Pattern Recognition}, pages 18051--18061.

\bibitem[{Meng et~al.(2021)Meng, He, Song, Song, Wu, Zhu, and Ermon}]{meng2021sdedit}
Chenlin Meng, Yutong He, Yang Song, Jiaming Song, Jiajun Wu, Jun-Yan Zhu, and Stefano Ermon. 2021.
\newblock Sdedit: Guided image synthesis and editing with stochastic differential equations.
\newblock \emph{arXiv preprint arXiv:2108.01073}.

\bibitem[{Muennighoff et~al.(2024)Muennighoff, Su, Wang, Yang, Wei, Yu, Singh, and Kiela}]{muennighoff2024generative}
Niklas Muennighoff, Hongjin Su, Liang Wang, Nan Yang, Furu Wei, Tao Yu, Amanpreet Singh, and Douwe Kiela. 2024.
\newblock Generative representational instruction tuning.
\newblock \emph{arXiv preprint arXiv:2402.09906}.

\bibitem[{Nichol et~al.(2022)Nichol, Dhariwal, Ramesh, Shyam, Mishkin, Mcgrew, Sutskever, and Chen}]{nichol2022glide}
Alexander~Quinn Nichol, Prafulla Dhariwal, Aditya Ramesh, Pranav Shyam, Pamela Mishkin, Bob Mcgrew, Ilya Sutskever, and Mark Chen. 2022.
\newblock Glide: Towards photorealistic image generation and editing with text-guided diffusion models.
\newblock In \emph{International Conference on Machine Learning}, pages 16784--16804. PMLR.

\bibitem[{Ramesh et~al.(2021)Ramesh, Pavlov, Goh, Gray, Voss, Radford, Chen, and Sutskever}]{ramesh2021zero}
Aditya Ramesh, Mikhail Pavlov, Gabriel Goh, Scott Gray, Chelsea Voss, Alec Radford, Mark Chen, and Ilya Sutskever. 2021.
\newblock Zero-shot text-to-image generation.
\newblock In \emph{International Conference on Machine Learning}, pages 8821--8831. PMLR.

\bibitem[{Rombach et~al.(2022)Rombach, Blattmann, Lorenz, Esser, and Ommer}]{rombach2022high}
Robin Rombach, Andreas Blattmann, Dominik Lorenz, Patrick Esser, and Bj{\"o}rn Ommer. 2022.
\newblock High-resolution image synthesis with latent diffusion models.
\newblock In \emph{Proceedings of the IEEE/CVF Conference on Computer Vision and Pattern Recognition}, pages 10684--10695.

\bibitem[{Rostamzadeh et~al.(2018)Rostamzadeh, Hosseini, Boquet, Stokowiec, Zhang, Jauvin, and Pal}]{rostamzadeh2018fashion}
Negar Rostamzadeh, Seyedarian Hosseini, Thomas Boquet, Wojciech Stokowiec, Ying Zhang, Christian Jauvin, and Chris Pal. 2018.
\newblock Fashion-gen: The generative fashion dataset and challenge.
\newblock \emph{arXiv preprint arXiv:1806.08317}.

\bibitem[{Ruiz et~al.(2023)Ruiz, Li, Jampani, Pritch, Rubinstein, and Aberman}]{ruiz2023dreambooth}
Nataniel Ruiz, Yuanzhen Li, Varun Jampani, Yael Pritch, Michael Rubinstein, and Kfir Aberman. 2023.
\newblock Dreambooth: Fine tuning text-to-image diffusion models for subject-driven generation.
\newblock In \emph{Proceedings of the IEEE/CVF Conference on Computer Vision and Pattern Recognition}, pages 22500--22510.

\bibitem[{Saharia et~al.(2022)Saharia, Chan, Saxena, Li, Whang, Denton, Ghasemipour, Gontijo~Lopes, Karagol~Ayan, Salimans et~al.}]{saharia2022photorealistic}
Chitwan Saharia, William Chan, Saurabh Saxena, Lala Li, Jay Whang, Emily~L Denton, Kamyar Ghasemipour, Raphael Gontijo~Lopes, Burcu Karagol~Ayan, Tim Salimans, et~al. 2022.
\newblock Photorealistic text-to-image diffusion models with deep language understanding.
\newblock \emph{Advances in Neural Information Processing Systems}, 35:36479--36494.

\bibitem[{Schwenk et~al.(2022)Schwenk, Khandelwal, Clark, Marino, and Mottaghi}]{schwenk2022okvqa}
Dustin Schwenk, Apoorv Khandelwal, Christopher Clark, Kenneth Marino, and Roozbeh Mottaghi. 2022.
\newblock A-okvqa: A benchmark for visual question answering using world knowledge.
\newblock In \emph{European Conference on Computer Vision}, pages 146--162. Springer.

\bibitem[{Shen et~al.(2023)Shen, Song, Tan, Li, Lu, and Zhuang}]{shen2023hugginggpt}
Yongliang Shen, Kaitao Song, Xu~Tan, Dongsheng Li, Weiming Lu, and Yueting Zhuang. 2023.
\newblock Hugginggpt: Solving ai tasks with chatgpt and its friends in huggingface.
\newblock \emph{arXiv preprint arXiv:2303.17580}.

\bibitem[{Shi et~al.(2023)Shi, Li, Zhang, Chen, and Wu}]{shi2023recon}
Guangyuan Shi, Qimai Li, Wenlong Zhang, Jiaxin Chen, and Xiao-Ming Wu. 2023.
\newblock Recon: Reducing conflicting gradients from the root for multi-task learning.
\newblock \emph{arXiv preprint arXiv:2302.11289}.

\bibitem[{Sohl-Dickstein et~al.(2015)Sohl-Dickstein, Weiss, Maheswaranathan, and Ganguli}]{sohl2015deep}
Jascha Sohl-Dickstein, Eric Weiss, Niru Maheswaranathan, and Surya Ganguli. 2015.
\newblock Deep unsupervised learning using nonequilibrium thermodynamics.
\newblock In \emph{International conference on machine learning}, pages 2256--2265. PMLR.

\bibitem[{Song et~al.(2020)Song, Meng, and Ermon}]{song2020denoising}
Jiaming Song, Chenlin Meng, and Stefano Ermon. 2020.
\newblock Denoising diffusion implicit models.
\newblock \emph{arXiv preprint arXiv:2010.02502}.

\bibitem[{Touvron et~al.(2023)Touvron, Lavril, Izacard, Martinet, Lachaux, Lacroix, Rozi{\`e}re, Goyal, Hambro, Azhar et~al.}]{touvron2023llama}
Hugo Touvron, Thibaut Lavril, Gautier Izacard, Xavier Martinet, Marie-Anne Lachaux, Timoth{\'e}e Lacroix, Baptiste Rozi{\`e}re, Naman Goyal, Eric Hambro, Faisal Azhar, et~al. 2023.
\newblock Llama: Open and efficient foundation language models.
\newblock \emph{arXiv preprint arXiv:2302.13971}.

\bibitem[{Ventura et~al.(2024)Ventura, Yang, Schmid, and Varol}]{ventura2024covr}
Lucas Ventura, Antoine Yang, Cordelia Schmid, and G{\"u}l Varol. 2024.
\newblock Covr: Learning composed video retrieval from web video captions.
\newblock In \emph{Proceedings of the AAAI Conference on Artificial Intelligence}, volume~38, pages 5270--5279.

\bibitem[{Wang et~al.(2022{\natexlab{a}})Wang, Yang, Men, Lin, Bai, Li, Ma, Zhou, Zhou, and Yang}]{wang2022ofa}
Peng Wang, An~Yang, Rui Men, Junyang Lin, Shuai Bai, Zhikang Li, Jianxin Ma, Chang Zhou, Jingren Zhou, and Hongxia Yang. 2022{\natexlab{a}}.
\newblock Ofa: Unifying architectures, tasks, and modalities through a simple sequence-to-sequence learning framework.
\newblock In \emph{International Conference on Machine Learning}, pages 23318--23340. PMLR.

\bibitem[{Wang et~al.(2022{\natexlab{b}})Wang, Bao, Dong, Bjorck, Peng, Liu, Aggarwal, Mohammed, Singhal, Som et~al.}]{wang2022image}
Wenhui Wang, Hangbo Bao, Li~Dong, Johan Bjorck, Zhiliang Peng, Qiang Liu, Kriti Aggarwal, Owais~Khan Mohammed, Saksham Singhal, Subhojit Som, et~al. 2022{\natexlab{b}}.
\newblock Image as a foreign language: Beit pretraining for all vision and vision-language tasks.
\newblock \emph{arXiv preprint arXiv:2208.10442}.

\bibitem[{Wen et~al.(2023)Wen, Zhang, Song, Wei, and Nie}]{wen2023target}
Haokun Wen, Xian Zhang, Xuemeng Song, Yinwei Wei, and Liqiang Nie. 2023.
\newblock Target-guided composed image retrieval.
\newblock In \emph{Proceedings of the 31st ACM International Conference on Multimedia}, pages 915--923.

\bibitem[{Wu et~al.(2023)Wu, Yin, Qi, Wang, Tang, and Duan}]{wu2023visual}
Chenfei Wu, Shengming Yin, Weizhen Qi, Xiaodong Wang, Zecheng Tang, and Nan Duan. 2023.
\newblock Visual chatgpt: Talking, drawing and editing with visual foundation models.
\newblock \emph{arXiv preprint arXiv:2303.04671}.

\bibitem[{Wu et~al.(2021)Wu, Gao, Guo, Al-Halah, Rennie, Grauman, and Feris}]{wu2021fashion}
Hui Wu, Yupeng Gao, Xiaoxiao Guo, Ziad Al-Halah, Steven Rennie, Kristen Grauman, and Rogerio Feris. 2021.
\newblock Fashion iq: A new dataset towards retrieving images by natural language feedback.
\newblock In \emph{Proceedings of the IEEE/CVF Conference on computer vision and pattern recognition}, pages 11307--11317.

\bibitem[{Xie et~al.(2023)Xie, Huang, Dong, Zhao, Dong, Zhang, Zhu, and Liang}]{xie2023gp}
Zhenyu Xie, Zaiyu Huang, Xin Dong, Fuwei Zhao, Haoye Dong, Xijin Zhang, Feida Zhu, and Xiaodan Liang. 2023.
\newblock Gp-vton: Towards general purpose virtual try-on via collaborative local-flow global-parsing learning.
\newblock In \emph{Proceedings of the IEEE/CVF Conference on Computer Vision and Pattern Recognition}, pages 23550--23559.

\bibitem[{Yang et~al.(2023{\natexlab{a}})Yang, Gu, Zhang, Zhang, Chen, Sun, Chen, and Wen}]{yang2023paint}
Binxin Yang, Shuyang Gu, Bo~Zhang, Ting Zhang, Xuejin Chen, Xiaoyan Sun, Dong Chen, and Fang Wen. 2023{\natexlab{a}}.
\newblock Paint by example: Exemplar-based image editing with diffusion models.
\newblock In \emph{Proceedings of the IEEE/CVF Conference on Computer Vision and Pattern Recognition}, pages 18381--18391.

\bibitem[{Yang et~al.(2020)Yang, Zhang, Jin, Liu, Wu, Tan, Xie, Wang, and Wang}]{yang2020fashion}
Xuewen Yang, Heming Zhang, Di~Jin, Yingru Liu, Chi-Hao Wu, Jianchao Tan, Dongliang Xie, Jue Wang, and Xin Wang. 2020.
\newblock Fashion captioning: Towards generating accurate descriptions with semantic rewards.
\newblock In \emph{Computer Vision--ECCV 2020: 16th European Conference, Glasgow, UK, August 23--28, 2020, Proceedings, Part XIII 16}, pages 1--17. Springer.

\bibitem[{Yang et~al.(2023{\natexlab{b}})Yang, Li, Wang, Lin, Azarnasab, Ahmed, Liu, Liu, Zeng, and Wang}]{yang2023mm}
Zhengyuan Yang, Linjie Li, Jianfeng Wang, Kevin Lin, Ehsan Azarnasab, Faisal Ahmed, Zicheng Liu, Ce~Liu, Michael Zeng, and Lijuan Wang. 2023{\natexlab{b}}.
\newblock Mm-react: Prompting chatgpt for multimodal reasoning and action.
\newblock \emph{arXiv preprint arXiv:2303.11381}.

\bibitem[{Ye et~al.(2023)Ye, Xu, Xu, Ye, Yan, Zhou, Wang, Hu, Shi, Shi et~al.}]{ye2023mplug}
Qinghao Ye, Haiyang Xu, Guohai Xu, Jiabo Ye, Ming Yan, Yiyang Zhou, Junyang Wang, Anwen Hu, Pengcheng Shi, Yaya Shi, et~al. 2023.
\newblock mplug-owl: Modularization empowers large language models with multimodality.
\newblock \emph{arXiv preprint arXiv:2304.14178}.

\bibitem[{Zang et~al.(2021)Zang, Liu, Wang, Song, Zhang, and Chen}]{zang2021photochat}
Xiaoxue Zang, Lijuan Liu, Maria Wang, Yang Song, Hao Zhang, and Jindong Chen. 2021.
\newblock Photochat: A human-human dialogue dataset with photo sharing behavior for joint image-text modeling.
\newblock In \emph{Proceedings of the 59th Annual Meeting of the Association for Computational Linguistics and the 11th International Joint Conference on Natural Language Processing (Volume 1: Long Papers)}, pages 6142--6152.

\bibitem[{Zhang et~al.(2022)Zhang, Liang, Zhang, Zhan, Wu, Lu, and Lam}]{zhang2022fine}
Haode Zhang, Haowen Liang, Yuwei Zhang, Li-Ming Zhan, Xiao-Ming Wu, Xiaolei Lu, and Albert Lam. 2022.
\newblock Fine-tuning pre-trained language models for few-shot intent detection: Supervised pre-training and isotropization.
\newblock In \emph{Proceedings of the 2022 Conference of the North American Chapter of the Association for Computational Linguistics: Human Language Technologies}, pages 532--542.

\bibitem[{Zhang et~al.(2023{\natexlab{a}})Zhang, Rao, and Agrawala}]{zhang2023adding}
Lvmin Zhang, Anyi Rao, and Maneesh Agrawala. 2023{\natexlab{a}}.
\newblock Adding conditional control to text-to-image diffusion models.
\newblock In \emph{Proceedings of the IEEE/CVF International Conference on Computer Vision}, pages 3836--3847.

\bibitem[{Zhang et~al.(2023{\natexlab{b}})Zhang, Han, Zhou, Hu, Yan, Lu, Li, Gao, and Qiao}]{zhang2023llama}
Renrui Zhang, Jiaming Han, Aojun Zhou, Xiangfei Hu, Shilin Yan, Pan Lu, Hongsheng Li, Peng Gao, and Yu~Qiao. 2023{\natexlab{b}}.
\newblock Llama-adapter: Efficient fine-tuning of language models with zero-init attention.
\newblock \emph{arXiv preprint arXiv:2303.16199}.

\bibitem[{Zhao et~al.(2024)Zhao, Liu, Liu, Shi, and Wu}]{zhao-etal-2024-easygen}
Xiangyu Zhao, Bo~Liu, Qijiong Liu, Guangyuan Shi, and Xiao-Ming Wu. 2024.
\newblock \href {https://aclanthology.org/2024.acl-long.74} {{E}asy{G}en: Easing multimodal generation with {B}i{D}iffuser and {LLM}s}.
\newblock In \emph{Proceedings of the 62nd Annual Meeting of the Association for Computational Linguistics (Volume 1: Long Papers)}, pages 1351--1370, Bangkok, Thailand. Association for Computational Linguistics.

\bibitem[{Zheng et~al.(2023)Zheng, Chiang, Sheng, Zhuang, Wu, Zhuang, Lin, Li, Li, Xing et~al.}]{zheng2023judging}
Lianmin Zheng, Wei-Lin Chiang, Ying Sheng, Siyuan Zhuang, Zhanghao Wu, Yonghao Zhuang, Zi~Lin, Zhuohan Li, Dacheng Li, Eric Xing, et~al. 2023.
\newblock Judging llm-as-a-judge with mt-bench and chatbot arena.
\newblock \emph{Advances in Neural Information Processing Systems}, 36:46595--46623.

\bibitem[{Zhu et~al.(2023{\natexlab{a}})Zhu, Chen, Shen, Li, and Elhoseiny}]{zhu2023minigpt}
Deyao Zhu, Jun Chen, Xiaoqian Shen, Xiang Li, and Mohamed Elhoseiny. 2023{\natexlab{a}}.
\newblock Minigpt-4: Enhancing vision-language understanding with advanced large language models.
\newblock \emph{arXiv preprint arXiv:2304.10592}.

\bibitem[{Zhu et~al.(2023{\natexlab{b}})Zhu, Wei, Zhao, Zhang, and Huang}]{zhu2023amc}
Hongguang Zhu, Yunchao Wei, Yao Zhao, Chunjie Zhang, and Shujuan Huang. 2023{\natexlab{b}}.
\newblock Amc: Adaptive multi-expert collaborative network for text-guided image retrieval.
\newblock \emph{ACM Transactions on Multimedia Computing, Communications and Applications}, 19(6):1--22.

\bibitem[{Zhuge et~al.(2021)Zhuge, Gao, Fan, Jin, Chen, Zhou, Qiu, and Shao}]{zhuge2021kaleido}
Mingchen Zhuge, Dehong Gao, Deng-Ping Fan, Linbo Jin, Ben Chen, Haoming Zhou, Minghui Qiu, and Ling Shao. 2021.
\newblock Kaleido-bert: Vision-language pre-training on fashion domain.
\newblock In \emph{Proceedings of the IEEE/CVF conference on computer vision and pattern recognition}, pages 12647--12657.

\end{thebibliography}

\appendix

\begin{table*}[t]
	\centering
	\fontsize{7}{8}\selectfont

	\begin{tabular}{l|l|c|c|c|c}
		\toprule
         Data types&Dataset&Size&Stage 1&Stage 2&Metrics\cr
          \midrule
         \multirow{2}{*}{CMR}&FashionGen~\citep{lin2014microsoft} & 260.5K & \CheckmarkBold & \CheckmarkBold & R@K \cr
          &Fashion200K~\citep{krishna2017visual} & 172K & \CheckmarkBold &  \XSolidBrush & - \cr
         \midrule
          CIR & Fashion-IQ~\cite{liu2023visual} & 18K & \XSolidBrush & \CheckmarkBold & R@K \cr
         \midrule
         \multirow{2}{*}{FIC}& FashionGen~\cite{liu2023visual} & 260.5K & \CheckmarkBold & \CheckmarkBold & BLEU,CIDEr,METEOR,ROUGE-L \cr
           &Fashion-IQ-Cap & 60K & \CheckmarkBold &  \XSolidBrush & - \cr
         \midrule
         \multirow{2}{*}{FIG}&  VITON-HD~\citep{goyal2017making} & 83K & \XSolidBrush & \CheckmarkBold & FID, KID \cr
         &MGD~\citep{schwenk2022okvqa} & 66K & \XSolidBrush & \CheckmarkBold & FID,KID,CLIP-S \cr
		\bottomrule
	\end{tabular}
        \caption{ Description of datasets used in two stages.}
        \label{VQA datasets}
\end{table*}

\begin{figure*}[ht]
	\begin{center}
	\includegraphics[width=0.98\textwidth]{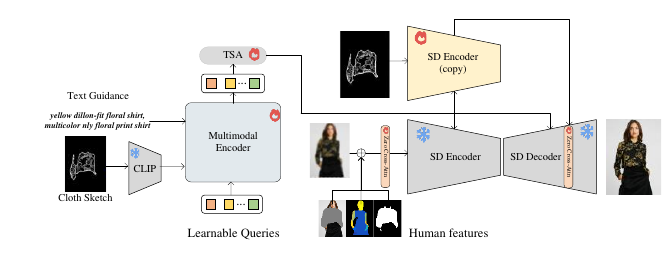}
        \caption{The architecture of \modelname{} for fine-tuning on the image editing task. Firstly, we supply the cloth sketch and text guidance to the multimodal encoder. Then, the diffusion model receives the output of the multimodal encoder, along with the cloth sketches and human features (i.e., agnostic-mask), to subsequently generate the desired images.
        }
	\label{vqa_tune}
    \end{center}
\end{figure*}

\section{Basics of Diffusion Models}

After the initial proposal of diffusion models by~\citep{sohl2015deep}, they have demonstrated remarkable capacity for generating high-quality and diverse data.
DDPM~\citep{ho2020denoising} connects diffusion and score matching models through a noise prediction formulation, while DDIM~\citep{song2020denoising} proposes an implicit generative model that generates deterministic samples from latent variables.

Given a data point sampled from a real data distribution $x_{0} \in q(x)$, during forward diffusion, $x_{0}$ is gradually “corrupted” at each step $t$ by adding Gaussian noise to the output of step $t$-$1$. It produces a sequence of noisy samples $\mathbf{x}_{1},\cdots,\mathbf{x}_{T}$. Then, diffusion models learn to reverse the process:
\begin{equation}
\begin{aligned}
    p(\mathbf{x}_{0:T}) &= p(\mathbf{x}_T)\prod_{t=1}^{T}p_{\theta}(\mathbf{x}_{t-1}|\mathbf{x}_t), \\ \quad p_{\theta}(\mathbf{x}_{t-1}|\mathbf{x}_{t}) &= 
    \mathcal{N}(\mathbf{x}_{t-1}; \mu_{t}(\mathbf{x}_{t},t),\sigma_t^{2}\mathbf{I}),
\end{aligned}
\end{equation}

where $p(\mathbf{x}_T)=\mathcal{N}(\mathbf{x}_T; 0,\mathbf{I})$ is  
the standard Gaussian distribution and $\mu_{t}(\cdot)$ is the parameterization of the predicted mean. Diffusion models are trained to maximize the marginal likelihood of the data $\mathbb{E} [\log p_{\theta}(\mathbf{x}_{0})]$, and the canonical objective is the variational lower bound of $\log p_{\theta}(\mathbf{x}_{0})$. 

\paragraph{Stable Diffusion Model.} 
Latent diffusion models (LDMs) operate in the latent space of a pre-trained autoencoder achieving higher computational efficiency while preserving the generation quality. Stable diffusion model is composed of an autoencoder with an encoder $\mathbb{E}$ and a decoder $\mathbb{D}$, a conditional U-Net denoising model $\bm{\epsilon}_{\theta}$, and a CLIP-based text encoder. With the fixed encoder $\mathbb{E}$, an input image $x$ is first transformed to a lower-dimensional latent space $z_{0} = \mathbb{E}(x)$. The decoder $\mathbb{D}$ performs the opposite operation, decoding $z_{0}$ into the pixel space. 
When considering a latent variable $z$ and its noisy counterpart $z_{t}$, which is obtained by incrementally adding noises to $z$ over $t$ steps, the latent diffusion models are designed to train the $\bm{\epsilon}_{\theta}(\cdot)$ to predict the added noise $\bm{\epsilon}$ using a standard mean squared error loss:

\begin{equation}
        \label{unconditional loss}
	\begin{aligned}
		\mathcal{L} := \mathbb{E}_{\bm z,{\bm{\epsilon}},t}[\Vert {\bm{\epsilon}} - \bm{\epsilon}_{\theta}(\mathbf{z}_{t},t) \Vert^{2}].
	\end{aligned}
\end{equation}

\paragraph{Multimodal Conditional Generation.} 
In the context of our current work, we have a particular focus on the pre-trained multimodal latent diffusion models. 
For a multimodal conditional generation, given a target image $\mathbf{x}_{0}$, the input condition $\mathbf{y}_{0}$ could contain different constraints. The aim is to model the conditional data distribution $q(\mathbf{x}_{0}|\mathbf{y}_{0})$, where $\mathbf{y}_{0}$ contains different modalities prompts.
The conditioning mechanism is implemented by first encoding conditional information, then the denoising network $\bm{\epsilon}_{\theta}$ conditions on $y_{0}$ via cross-attention. The label $y_{0}$ in a class-conditional diffusion model $\bm{\epsilon}_{\theta}(x_{t}|y_{0})$ is replaced with a null label $\emptyset$ with a fixed probability during training. 



\section{Implementation Details}
\label{implementation}


\paragraph{LLM} During the first phase, due to the flexibility brought by the modular architectural design of BLIP-2, we are able to adapt the model to a broad spectrum of LLMs. In order to effectively utilize the capabilities of the existing MLLM models, we adopted LLaVA-1.5 as the LLM module of the model. Technically, we leverage LoRA to enable a small subset of parameters within \modelname{} to be updated concurrently with two layers of adapter during this phase. Specifically, the lora rank is 128 and lora alpha is 256. We utilize the AdamW optimizer with $\beta_{0}$ = 0.9, $\beta_{1}$ = 0.99, and weight decay of 0. The LLMs are trained with a cosine learning rate of 2e-5 and a warmup rate of 0.03.  We use a batch size of 32 for the tuned LLMs. 

\paragraph{Diffusion Module} We inherit the autoencoder and the denoising U-Net of the Stable Diffusion v1.4. The weights of the U-Net from Paint-by-Example are used to initialize our denoising U-Net. To achieve more refined person texture, a VAE fine-tuned on the VITONHD dataset from StableVITON is utilized. We train the model using an AdamW optimizer with a fixed learning rate of 1e-4 for 360k iterations, employing a batch size of 32. For inference, we employ the pseudo linear multi-step sampler, with the number of sampling steps set to 50.

\begin{figure}[ht]
	\begin{center}
	\includegraphics[width=0.48\textwidth]{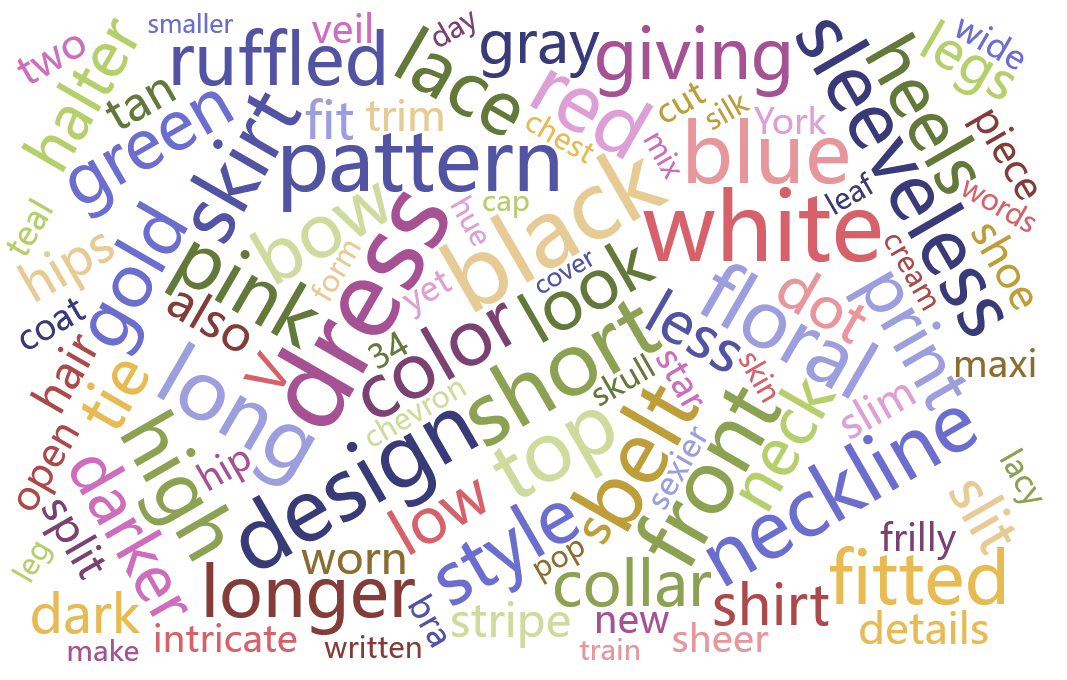}
        \caption{Vocabulary of the frequent words scaled by frequency for dresses.
        }
	\label{word_cloud}
    \end{center}
\end{figure}
\section{Datasets}
\label{dataset detail}

\begin{figure}[]
	\begin{center}
	\includegraphics[width=0.48\textwidth]{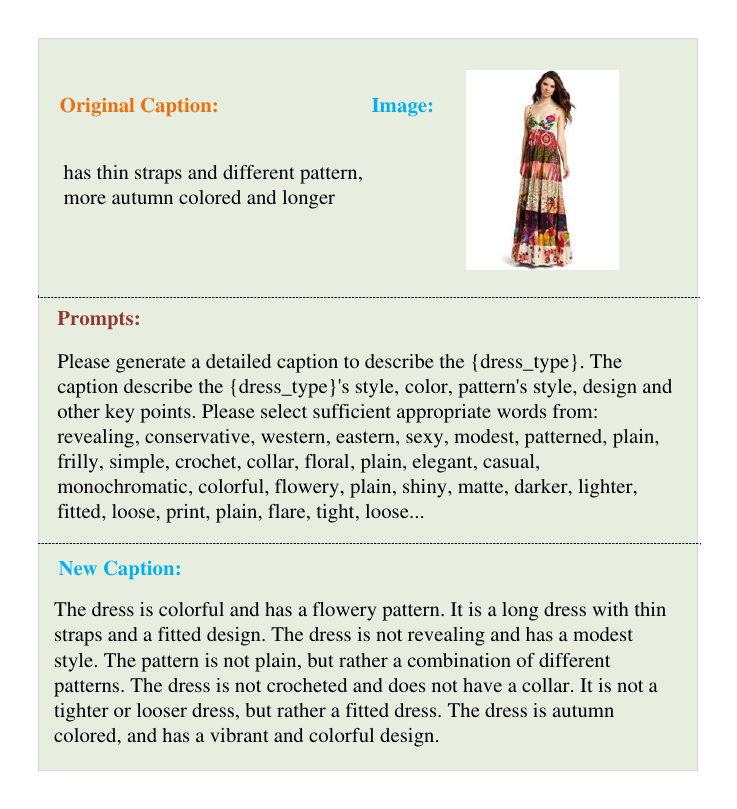}
        \caption{Illustration of Instruction-Following Data. The top section displays an image alongside its original captions from Fashion-IQ dataset. The bottom section presents detailed captions generated by LLaVA-1.5. The original captions are not prompts for generation but are provided for comparison with the newly generated caption.
        }
	\label{llavagen}
    \end{center}
\end{figure}

\begin{figure*}[ht]
	\begin{center}
	\includegraphics[width=0.98\textwidth]{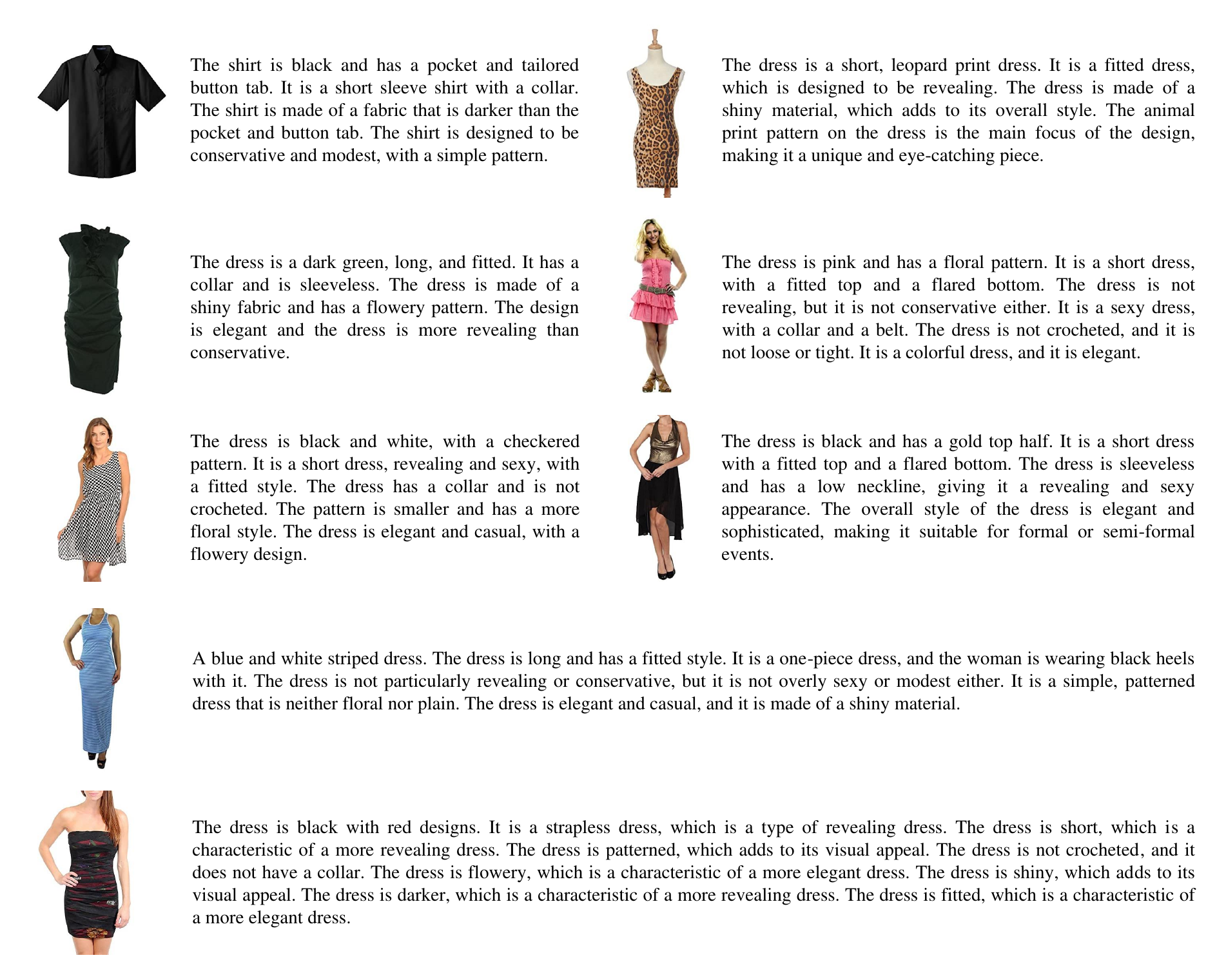}
        \caption{ 
        Caption generation results using our method with images from the Fashion-IQ dataset.
        }
	\label{fashion_caption}
    \end{center}
\end{figure*}

For fashion image captioning tasks, we utilize the FashionGen~\citep{zang2021photochat} dataset. Additionally, to enhance our model's capability in the CIR task, which involves the ability to retrieve captions for target images, we have annotated images from the training set of Fashion-IQ. Recognizing that manually annotating all the images would be time-consuming and resource-intensive, we draw inspiration from the success of recent MLLM models such as LLaVA in text-annotation tasks, and propose leveraging LLaVA 1.5 (13B) to semi-automatically annotate the dataset. We perform word lemmatization to reduce each word to its root form. Such pre-processing stage is crucial for the Fashion-IQ dataset, as the captions do not describe a single garment but instead express the properties to modify in a given image to match its target. As shown in Fig.~\ref{word_cloud}, by analysis of the captions in Fashion-IQ, we extracted key words that describe clothing information such as color, sleeve, pattern, lace, etc., as prompts for MLLM (LLaVA 1.5). We then instructed the model to generate the corresponding captions referencing words that match the image features, as shown in Fig.~\ref{llavagen}. After this process, we got the captions for Fashion-IQ dataset. The trained \modelname{} from this dataset (Fashion-IQ-cap) can generate captions for images in the evaluation set of Fashion-IQ to assist in the CIR task. More results can be seen in Fig.~\ref{fashion_caption}.

\section{Instruction Formats}
\label{instructions for llm}

Due to the disparity in caption styles across different fashion datasets, we employ diverse instructions to fine-tune the LLM, enabling it to generate captions of varying styles. Specifically, the Fashion200K dataset inclines towards providing brief descriptions, the FashionGen dataset is prone to offering professional captions, and in Fashion-IQ-cap, the captions are detailed. Consequently, we have designed distinct instructions for different datasets and tasks, as illustrated in Table~\ref{instruction templates}. 


\begin{table*}[t]
	\centering
	\fontsize{8}{8}\selectfont
	\begin{tabular}{l|l}
		\toprule
          \textbf{Dataset}&\textbf{Instruction}\cr
          \midrule
         \textbf{Fashion200K}&USER:<image>+Short description. Assistant:\cr
         \midrule
         \textbf{FashionGen} & USER:<image>+Write a detail and professional description for the cloth. Assistant:\cr
         \midrule
         \textbf{Fashion-IQ-cap}&USER:<image>+Describe the cloth's style, color, design... and other key points. Assistant:\cr
		\bottomrule
	\end{tabular}
        \caption{Examples of task instruction templates. 
        }
        \label{instruction templates}
\end{table*}

\begin{table*}[ht]
	\centering
	\fontsize{7}{8}\selectfont

	\begin{tabular}{c|c|c|c|c|c|c|c}
		\toprule
         Model Types&Task Domain&Model&Main Structure&XMR&CIR&\begin{tabular}[c]{@{}c@{}}Text \\ Generation\end{tabular}&\begin{tabular}[c]{@{}c@{}}Image \\ Generation\end{tabular}\cr
          \midrule
         \multirow{2}{*}{ Cross-modal Retrieval}&General& CLIP (2021) & Dual-stream Transfomer & \CheckmarkBold & \XSolidBrush & \XSolidBrush & \XSolidBrush \cr
          &Fashion& FashionBERT (2020) & Single-stream Transfomer & \CheckmarkBold & \XSolidBrush & \XSolidBrush & \XSolidBrush \cr
         \midrule
          {Multimodal LLM} & General & LLaVA (2023) & CLIP, LLM & \XSolidBrush & \XSolidBrush & \CheckmarkBold  & \XSolidBrush \cr
          \midrule
        {Composed Image Retrieval} & General & SPRC (2024) & CLIP, Qformer & \XSolidBrush & \CheckmarkBold & \XSolidBrush  & \XSolidBrush \cr
         \midrule
         \multirow{2}{*}{Conditional Diffusion}& General & ControlNet (2023) & Stable diffusion & \XSolidBrush & \XSolidBrush & \XSolidBrush & \CheckmarkBold \cr
           &Fashion & StableVITON (2023) & Stable diffusion &  \XSolidBrush & \XSolidBrush & \XSolidBrush & \CheckmarkBold \cr
         \midrule
         \multirow{3}{*}{Unified Model}& General & NExT-GPT (2023) & ImageBind, LLM, Diffusion &  \XSolidBrush & \XSolidBrush & \CheckmarkBold & \CheckmarkBold \cr
         &Fashion & FAME-ViL (2023) & Dual-stream Transfomer & \CheckmarkBold & \CheckmarkBold & \CheckmarkBold  & \XSolidBrush \cr
         & General & BLIP2 (2023) & CLIP, Qformer, LLM & \CheckmarkBold & \XSolidBrush & \CheckmarkBold  & \XSolidBrush \cr
         \midrule
         \bf{Unified Model (Ours)}& Fashion & UniFashion & CLIP, Qformer, LLM, Diffusion &  \CheckmarkBold & \CheckmarkBold & \CheckmarkBold & \CheckmarkBold \cr
		\bottomrule
	\end{tabular}
        \caption{Comparison of different multimodal models.  \textbf{XMR}: Cross-modal retrieval tasks; \textbf{CIR}: Compoesd image retrieval task.}
        \label{Baselines}
\end{table*}

\section{Visual Results}
\label{fashion_design}

\begin{figure*}[ht]
	\begin{center}
	\includegraphics[width=0.95\textwidth]{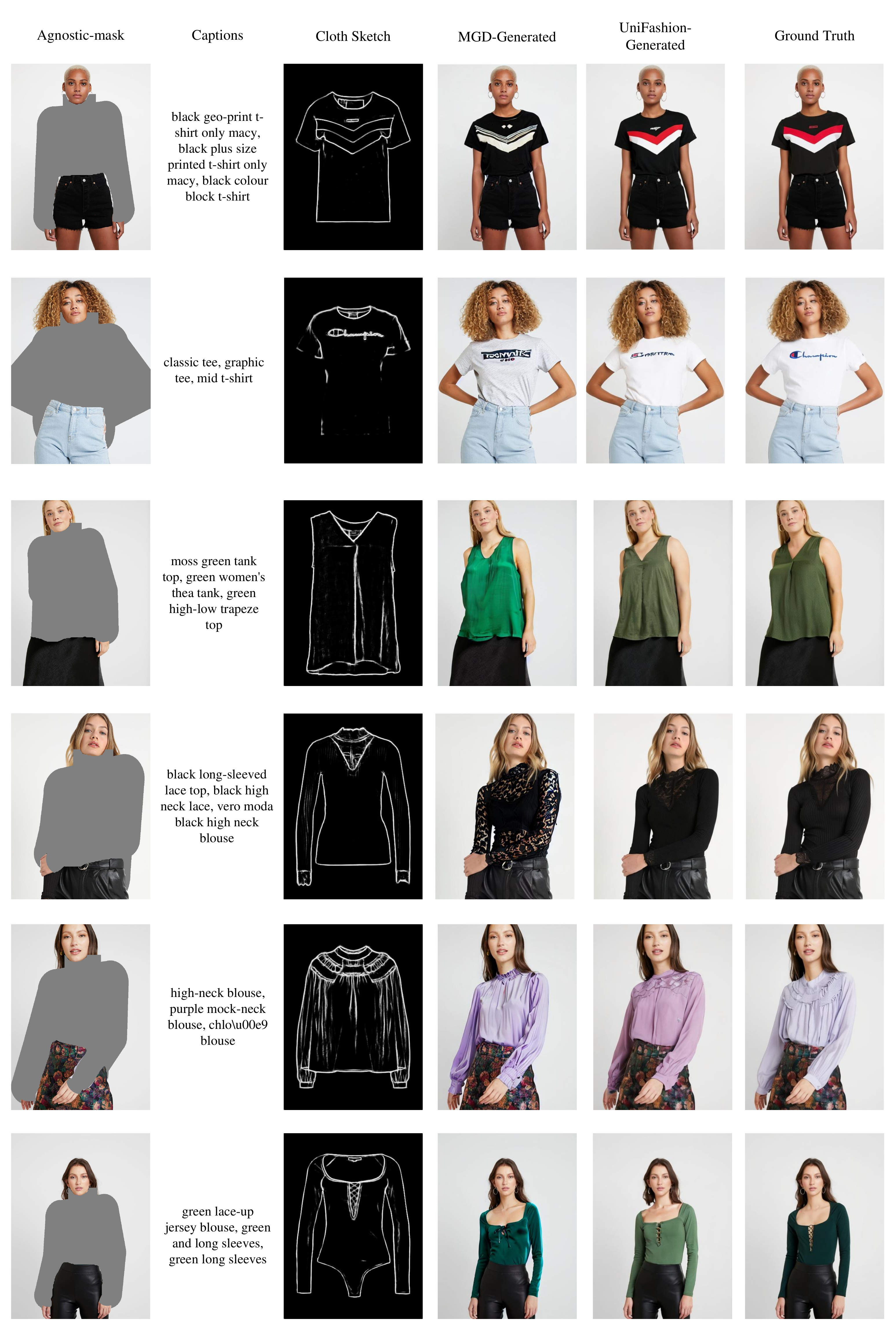}
        \caption{Qualitative comparison on VITON-HD paired test set. From left to right: agnostic-mask image, caption, cloth sketch, MGD-generated image, \modelname{} (ours)-generated image and ground truth. Our method is capable of generating images that align more accurately with the given captions and cloth sketch. For optimal viewing, please zoom in.}
	\label{paired_example}
    \end{center}
\end{figure*}

\begin{figure*}[ht]
	\begin{center}
	\includegraphics[width=0.98\textwidth]{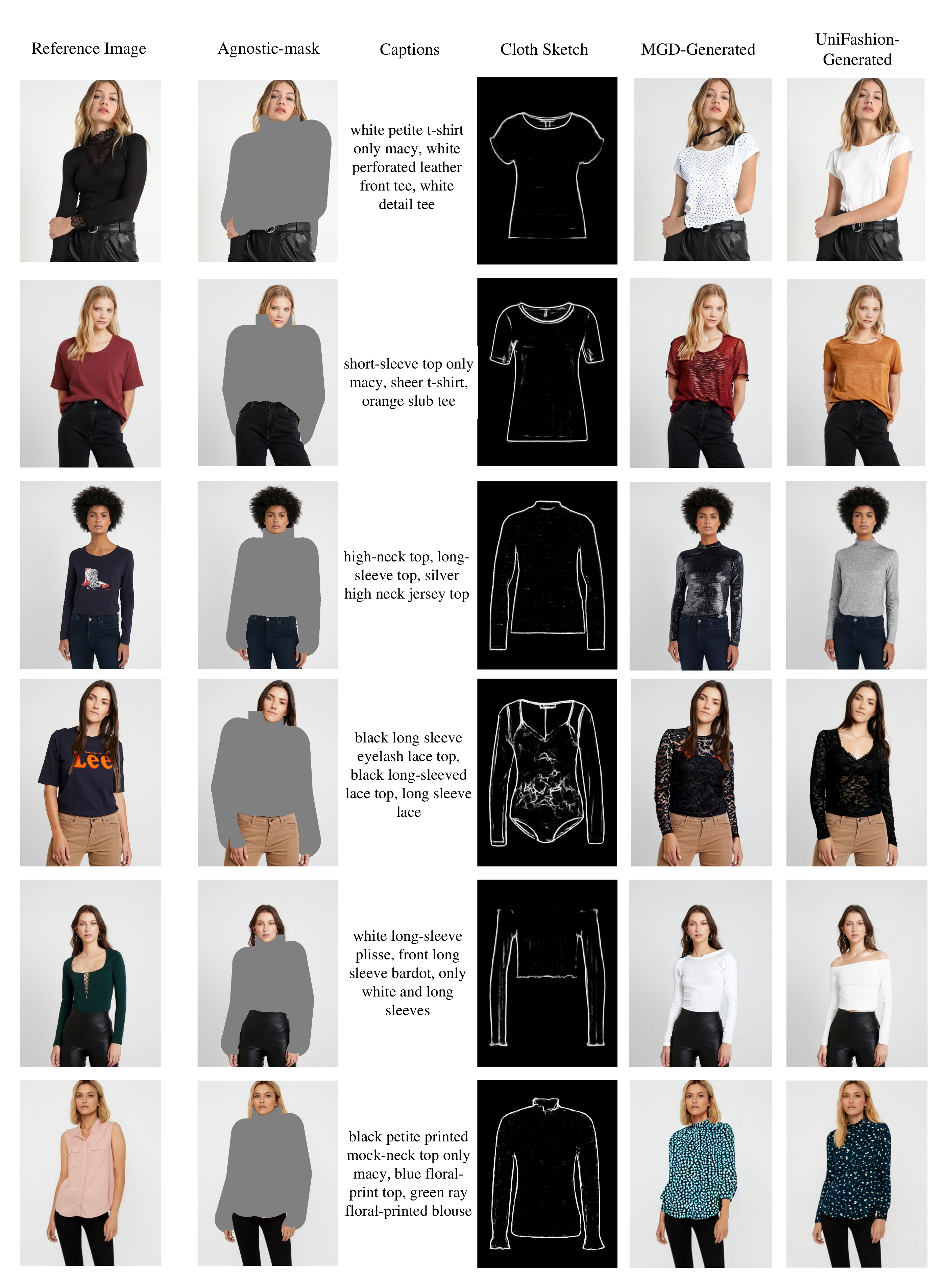}
        \caption{Qualitative comparison on VITON-HD unpaired test set. From left to right: original image, agnostic-mask image, captions, MGD input sketch, MGD-generated image, \modelname{} input sketch and \modelname{} (ours)-generated image. Our model is capable of generating images that align more accurately with the provided captions and cloth sketch. For optimal viewing, please zoom in.}
	\label{fashion_example}
    \end{center}
\end{figure*}

Figure~\ref{vqa_tune} illustrates the architecture of \modelname{} for fine-tuning on the image editing task. Initially, we input the cloth sketch and text guidance into the multimodal encoder. The diffusion model then receives the output from the multimodal encoder, along with the cloth sketches and human features (such as the agnostic mask), to generate the desired images. We compare \modelname{} with the MGD~\citep{baldrati2023multimodal} model for this task. In Fig.~\ref{paired_example}, we compare the images generated by our approach with the competitor in the VITON-HD~\citep{choi2021viton} paired setting. In Fig.~\ref{fashion_example}, we show the generation effects of \modelname{} in the VITON-HD unpaired setting. Our method, unlike the MGD method that employs a warping module to generate input sketches, directly uses in-shop garment sketches and is capable of generating images that align more accurately with the provided captions and cloth sketches.

\label{sec:appendix}

\end{document}